\documentclass[11pt,letterpaper]{article}
\usepackage[T1]{fontenc}
\usepackage{lmodern}
\usepackage{graphicx}
\usepackage{mathtools,amssymb}
\usepackage{xcolor}
\usepackage[numbers,sort&compress]{natbib}
\usepackage[hyphens]{url}
\usepackage[font=small,labelfont=bf,labelsep=period]{caption}
\usepackage{enumitem}
\usepackage{listings}
\usepackage{booktabs,array,colortbl,tabularx,multirow}
\usepackage{tcolorbox}
\usepackage{placeins}
\usepackage[colorlinks=true,allcolors=linkblue,breaklinks]{hyperref}
\usepackage[capitalize]{cleveref}

\definecolor{linkblue}{RGB}{35,75,120}
\definecolor{oursblue}{RGB}{230,239,248}
\definecolor{stageurbanA}{HTML}{FCF1EB}
\definecolor{stageurbanB}{HTML}{EFF5FB}
\definecolor{boxblue}{RGB}{43,102,173}
\definecolor{boxtitlegray}{RGB}{96,96,96}

\setlist[itemize]{leftmargin=*,itemsep=3pt,topsep=5pt}
\newcommand{\sftcount}{1,008}
\newcommand{\oursrow}{\rowcolor{oursblue}}
\newcommand{\oursrule}{%
  \arrayrulecolor{black!65}\specialrule{0.8pt}{1.5pt}{0pt}%
  \arrayrulecolor{black}}
\newcommand{\mbemph}[1]{{\normalfont\bfseries\itshape #1}}
\newcommand{\mbhead}[1]{{\normalfont\bfseries #1}}
\newtcolorbox{methodbox}[1]{
  arc=3pt,boxrule=0.7pt,
  colframe=boxtitlegray,colback=white,
  colbacktitle=boxtitlegray,coltitle=white,
  fonttitle=\normalsize\bfseries,title={#1},
  toptitle=3pt,bottomtitle=3pt,lefttitle=8pt,righttitle=8pt,
  halign title=center,left=0pt,right=0pt,top=3pt,bottom=4pt,
  boxsep=0pt,width=\linewidth,halign=left,
  before skip=\medskipamount,after skip=\medskipamount
}

\newcommand{\paperrevcolor}{}

\DeclareRobustCommand{\reviewrev}[1]{#1}

\DeclareRobustCommand{\citationrev}[1]{#1}

\DeclareRobustCommand{\joinrev}[1]{#1}
\newenvironment{paperrevision}{\par\begingroup}{\par\endgroup}
\newenvironment{reviewrevision}{\par\begingroup}{\par\endgroup}

\title{Gravity and Flight: Learning to Navigate Fine-Grained Remote Sensing Scenes}
\author{%
  \parbox{0.96\textwidth}{\centering
    \large
    Yuheng Zong\textsuperscript{1},
    Minghua Wang\textsuperscript{1},
    Xin Zhao\textsuperscript{1}\\[3pt]
    Xiaobin Liu\textsuperscript{2},
    Yongjian Deng\textsuperscript{3},
    Chanho Eom\textsuperscript{4}\\[9pt]
    \footnotesize
    \textsuperscript{1}The Institute of Robotics and Automatic Information System (IRAIS),
    the Tianjin Key Laboratory of Intelligent Robotics (tjKLIR),
    Nankai University, Tianjin, 300071, China\\[5pt]
    \textsuperscript{2}The College of Artificial Intelligence,
    Nankai University, Tianjin, 300071, China\\[5pt]
    \textsuperscript{3}The Tianjin Key Laboratory of Intelligent Robotics,
    the Academy for Advanced Interdisciplinary Studies, and the College of Artificial Intelligence,
    Nankai University, Tianjin, 300071, China\\[5pt]
    \textsuperscript{4}The Department of Imaging Science and Arts \& Metaverse Convergence,
    Chung-Ang University, South Korea\\[3pt]
    \href{mailto:cheom@cau.ac.kr}{\texttt{cheom@cau.ac.kr}}
  }}
\date{}

\hypersetup{
  pdftitle={Gravity and Flight: Learning to Navigate Fine-Grained Remote Sensing Scenes},
  pdfauthor={Yuheng Zong, Minghua Wang, Xin Zhao, Xiaobin Liu, Yongjian Deng, Chanho Eom}
}

\begin{document}
\maketitle
\begin{abstract}
\begin{paperrevision}
\noindent
Vision-language models are becoming increasingly capable, yet their reasoning
over complex remote sensing scenes can remain coarse-grained and
insufficiently grounded in task-relevant visual evidence. To improve remote
sensing scene specialization with limited supervision, we propose Gravity and
Flight (G\&F), a framework that separates visual and reasoning verification
while coupling policy updates to support faithful reasoning based on
fine-grained visual observation. We realize G\&F through two-stage
post-training comprising Evidence-of-Thought supervised fine-tuning (EoT-SFT)
and Relational Transfer Policy Optimization (RTPO). EoT-SFT teaches the model
to describe visual details, analyze task-relevant evidence, and draw supported
conclusions. RTPO then refines reasoning on scene-matched image pairs with
fine-grained visual differences under a shared task, generating responses
independently for each image. In its rubric-based verification, Gravity checks
visual fidelity and description coverage, while Flight takes the descriptions
as premises to assess evidence use and relations between paired judgments.
Verification and task-correctness scores yield a shared pair-level advantage
for policy updates from both responses. Using 1,008 demonstrations and 500
training pairs for harbor specialization, G\&F achieves a mean score of 69.12
across six evaluation sets, exceeding the backbone and supervised
initialization by 9.76 and 4.56 points, respectively. Matched controls show
gains over single-image and joint-input RL, while removing either verification
view reduces performance. To assess applicability across scene domains, we
apply G\&F to urban specialization, obtaining gains over both the base model
and supervised initialization across three evaluation subsets.
\textbf{\textit{For a more detailed project overview and ongoing updates, please visit
\href{https://github.com/Z0ngL1ng/Gravity-and-Flight}{https://github.com/Z0ngL1ng/Gravity-and-Flight}.}}
\end{paperrevision}
\end{abstract}

\begin{figure}[tbp]
  \centering
  \includegraphics[width=\linewidth]{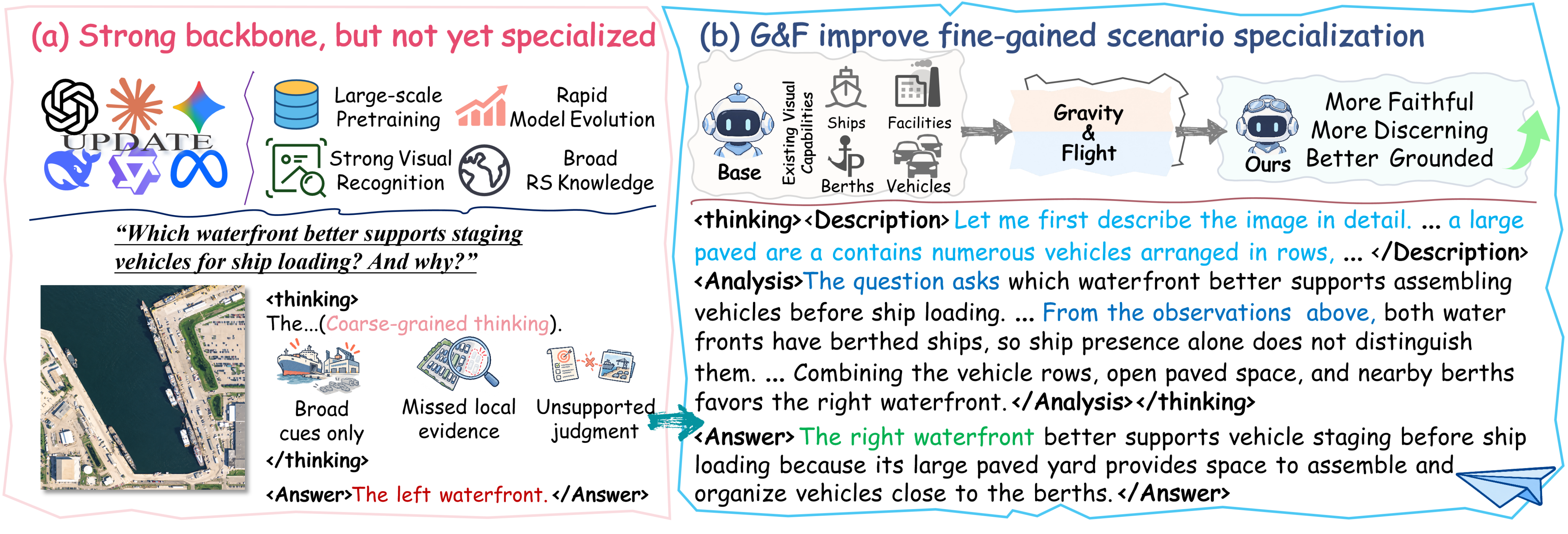}
  \caption{Motivation for G\&F. A harbor example contrasts
  recognizing scene objects with using task-relevant local evidence
  to judge vehicle-staging suitability.}
  \label{fig:motivation}
\end{figure}

\section{Introduction}

Vision-language models (VLMs) have developed strong recognition and
instruction-following capabilities~\cite{llava,cambrian}, with growing
applications to remote sensing~\cite{geochat,earthdial}. Beyond identifying
objects and scene categories, understanding complex remote sensing scenes
requires jointly interpreting local attributes, spatial relationships, and
task-specific conditions~\cite{finedefics,xlrsbench}. However, VLMs can
still produce coarse-grained analyses that overlook task-critical details or
draw conclusions beyond what the observed evidence supports~\cite{papo}.

As shown in \cref{fig:motivation}, the backbone recognizes vessels and
industrial buildings in the harbor scene, but these coarse-grained
observations alone cannot establish which waterfront better supports vehicle
staging before ship loading. The judgment requires inspecting vehicle rows,
usable paved space, and access to nearby berths, then explaining how their
arrangement supports vehicle assembly.

Learning such evidence-grounded judgments is particularly challenging when
high-quality annotated examples for scene specialization are scarce.
Separate assessments of visual fidelity and evidence use complement
final-answer scoring with feedback on the observations and reasoning behind
each answer. We use explicit rubrics to turn these assessments into training
scores and make fuller \mbox{use of each example}~\cite{rar}.

To improve remote sensing scene specialization through fine-grained visual
observation and faithful reasoning, we propose Gravity and Flight (G\&F), a
two-stage post-training framework. Evidence-of-Thought supervised
fine-tuning (EoT-SFT) initializes the model to distinguish observed scene
facts from their task-dependent interpretation. Its demonstrations record
local details and visibility limits in the description, then use relevant
observations in the analysis to support the conclusion.

Relational Transfer Policy Optimization (RTPO) organizes training around
scene-matched image pairs with fine-grained visual differences under a
shared task. Responses are sampled independently for each image and matched
into candidate pairs. Gravity checks visual claims and description coverage
against image evidence, while Flight assesses evidence use and the relation
between paired judgments conditional on the generated descriptions. Their
rubric scores and task correctness form a reward for each candidate pair.
Normalizing pair rewards within each training unit yields shared pair-level
advantages that couple the policy updates for the two responses in each
pair.

Across six harbor evaluation sets, G\&F achieves a higher mean score than
the backbone, supervised initialization, and matched RL controls. Removing
either verification mechanism reduces performance. Token-level analysis of six
sampled cases further shows more complete coverage of required visual facts
and better-supported judgments in \mbox{shorter responses}.

To assess applicability beyond harbors, we apply both stages to urban specialization with the same rubric dimensions and reward weights, obtaining
gains over both the base model and supervised initialization across three
evaluation subsets.

Our contributions are summarized as follows:

\begin{itemize}
    \item We introduce \textbf{G\&F}, a two-stage post-training framework for remote sensing
    scene specialization through fine-grained visual observation and
    faithful reasoning.

    \item We develop \textbf{RTPO}, which combines decoupled rubric-based verification with a
    \textbf{shared pair-level advantage} to couple policy updates from independently
    \mbox{generated responses}.

    \item We demonstrate gains over both the base model and supervised initialization
    in \textbf{harbor and urban specialization}, with ablations, matched controls,
    and token-level evidence from the harbor setting.
\end{itemize}

\section{Related Work}

\paragraph{Remote Sensing Vision-Language Models.}
Vision-language pretraining produces transferable visual
representations~\cite{clip,remoteclip}, which remote sensing VLMs
adapt through domain-specific instruction tuning
for grounded interaction and multi-source
understanding~\cite{geochat,earthdial,lhrsbot}, as well as detailed
object and relation descriptions~\cite{skysensegpt}.
FBA develops prerequisite remote-sensing and cross-domain
capabilities before evidence-grounded scene tuning~\cite{fba}.
Complementary approaches improve access to local evidence
through iterative inspection~\cite{rseot} and active cropping
or search~\cite{zoomearth,geovis}.
However, even when a crop resolves an object's identity, its function
may remain uncertain. G\&F checks the description against the image
and assesses whether those observations support the model's conclusion
under the stated task conditions.

\paragraph{Relational Supervision for Fine-Grained Reasoning.}
Image relations can make subtle visual distinctions explicit
through comparison~\cite{winoground,imgdiff}.
EqSim links semantic changes to image-text
similarity~\cite{eqsim}, while Fine-R1 uses within-category
and cross-category relations to improve fine-grained
recognition~\cite{finer1}.
MiCo builds triplets from two augmented views of one image and a
similar third image, rewarding same/different judgments
on joint inputs~\cite{mico}. Visually distinct scenes, however, can
support the same answer to a shared question. G\&F assesses whether
the paired descriptions warrant agreement or change in the independently
generated judgments.

\paragraph{Visual Faithfulness and Reasoning Rewards.}
Visual faithfulness studies diagnose unsupported object claims
and image-conditioned reasoning errors~\cite{pope,hallusionbench},
while fine-grained correctional feedback improves
alignment~\cite{rlhfv}. These concerns also motivate rewards for perceptual
consistency~\cite{perceptionr1} and intermediate visual
faithfulness~\cite{geosolver}.
\reviewrev{Vision-SR1 tests whether generated descriptions support
subsequent text-only reasoning~\cite{visionsr1}.
RAPID optimizes query-relevant captions using downstream answer
correctness from an external text reasoner~\cite{rapid}.
LLaVA-Critic and Prometheus-Vision evaluate multimodal response
quality~\cite{llavacritic,prometheusvision}.
Rubrics as Rewards uses instance-specific criteria as policy
feedback~\cite{rar}, while remote sensing models also optimize
answer rewards~\cite{rsthinker,geozero}.}
G\&F evaluates relations across task-matched scenes and uses
a shared pair advantage to couple the policy updates.

\section{Method}
\label{sec:method}

\subsection{Overview}
\label{sec:problem}
\label{sec:overview}

\begin{figure}[t]
  \centering
  \includegraphics[width=\textwidth]{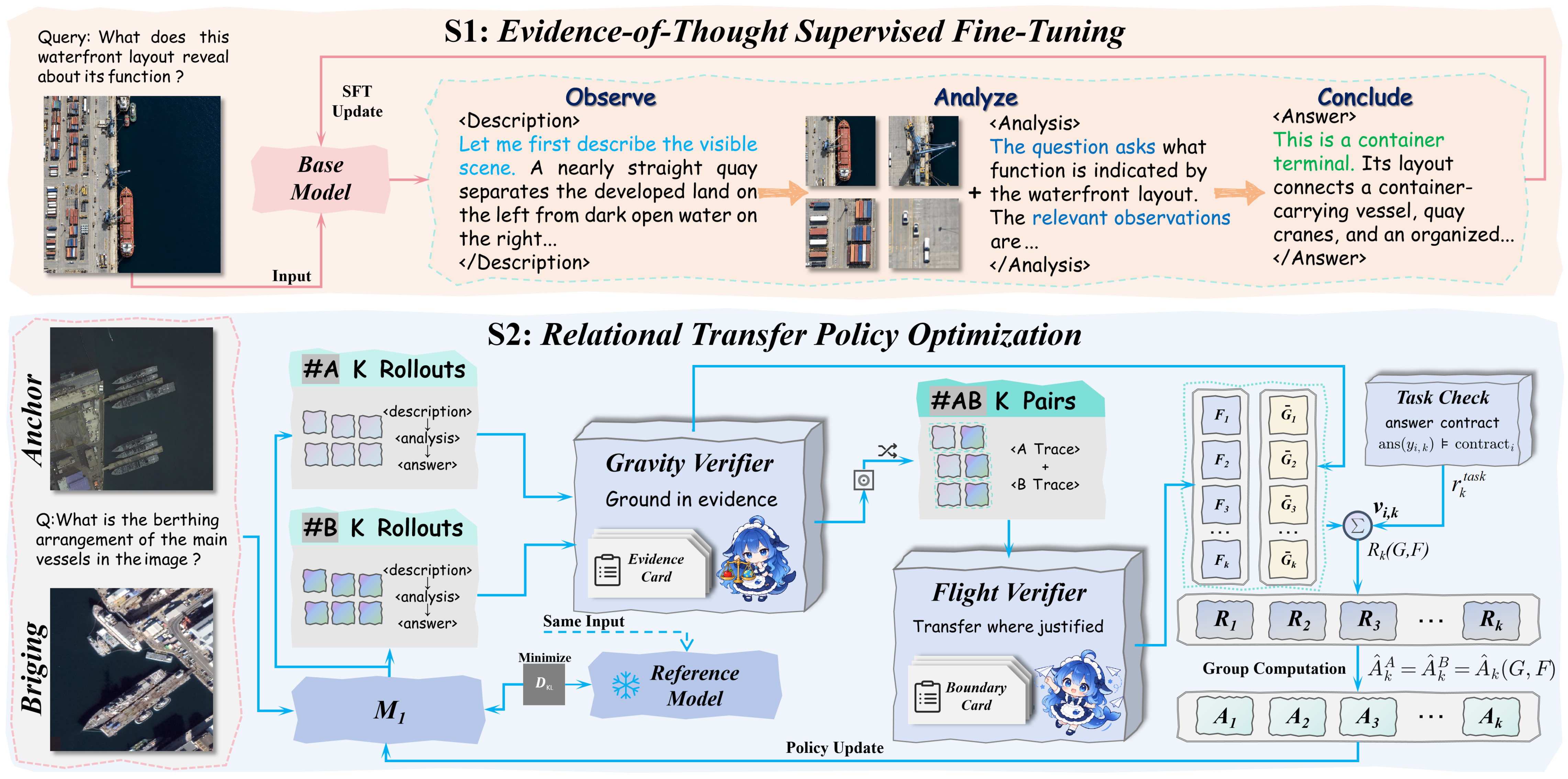}
  \caption{Overview of G\&F. EoT-SFT provides supervised initialization.
  RTPO jointly evaluates independently generated responses with Gravity
  and Flight, then updates the policy using a shared pair-level advantage.}
  \label{fig:framework}
\end{figure}

\reviewrev{ We adapt a pretrained VLM parameterized by $\theta$,
denoted as $\pi_\theta(y\mid x,q)$, where $x$ is the
input image, $q$ is the public task instruction, and
$y$ is the generated response comprising a scene
description, task analysis, and final answer.
As illustrated in \cref{fig:framework}, G\&F follows a two-stage framework. EoT-SFT establishes an
observation-to-judgment process, while RTPO further refines it through relational feedback from task-matched scenes.}

\subsection{EoT Supervised Initialization}
\label{sec:eot}
\label{sec:harbor-eot-data}
\label{sec:eot-sft}

\begin{reviewrevision}
EoT-SFT uses supervised reasoning traces~\cite{distillstep,mmcot}
to establish detailed observations before task-specific analysis.
The supervised targets distinguish what the image shows from how
those observations support an answer.

We construct the demonstrations through three steps of EoT
distillation, comprising scene observation, task interpretation, and
consistency review. In remote sensing scenes, numerous objects and facilities coexist within complex spatial layouts. Scene observation first builds an Evidence Card,
a structured image-level record (\cref{sec:anchor-bridge}), covering principal subjects,
salient secondary objects, local attributes, counts, spatial relations,
and visibility limits. Each entry includes a textual region reference,
visibility, confidence, and coverage priority. This image-level
record supplies evidence for different tasks.
In spatially complex remote sensing scenes, linking observations
to their corresponding objects or regions helps determine which entities exhibit a given attribute, participate in a spatial relation, or qualify for inclusion in a count. For counting questions, these associations identify objects
within the requested spatial scope. For functional questions,
they link facilities to nearby access routes and visible activity.
Visibility limits further distinguish unresolved attributes
from observed absences, ensuring that task-specific judgments
remain grounded in visible evidence without overstating
uncertain details.
Task interpretation then identifies the target and spatial scope in the
public instruction $q$, selects relevant observations, and applies
its conditions or rules. Alternative interpretations are examined
through the visible details that distinguish them. For the vehicle-staging
task in \cref{fig:motivation}, the analysis considers vehicle rows,
usable paved space, and proximity to berths. Ships are present at
both waterfronts, so suitability depends on whether the local arrangement
provides space for vehicle assembly and \mbox{access to ships for loading}.

Finally, consistency review checks the demonstrations for visual
support~\cite{hallucidoctor}, evidence use in the analysis, and agreement
of the answer with the question and preceding reasoning.
When a task-critical detail is obscured or unresolved, the target
response records that uncertainty and retains a broader conclusion
whenever the remaining visible evidence supports one.
The corpus contains \sftcount{} demonstrations over 888 images, with
7,745 observation annotations, 1,411 uncertainty annotations, and
1,698 assessments of alternative interpretations. The demonstrations
cover local attributes, counting, spatial and connection reasoning,
functional interpretation, and judgments governed by public conditions.
Construction and review details are in \cref{app:data}.

Each example pairs an image $x$ and public instruction $q$ with a
target response $y^*$. The description presents scene observations,
the analysis selects and applies task-relevant evidence, and the
answer states the resulting judgment. Questions about the same image
share a description, while their analyses and answers follow the task.
The target response is structured as
\end{reviewrevision}

\begin{equation}
y^*=\underbrace{\mathrm{description}\rightarrow\mathrm{analysis}}_{\text{EoT trace}}
\;\oplus\;\underbrace{\mathrm{answer}}_{\text{final}}.
\label{eq:eot-format}
\end{equation}

Here, $\oplus$ appends the final answer to the EoT trace.

\begin{methodbox}{Response Template}
\begin{lstlisting}
INPUT image x and public instruction q
<think>
  <description>
    Scene overview and salient objects.
    Attributes, ownership, spatial relations.
    Visibility, occlusion, and crop limits.
  </description>
  <analysis>
    Identify target, scope, and conditions.
    Select relevant description facts.
    Apply task rules within evidence limits.
  </analysis>
</think>
<answer>(*@\mbemph{Supported final answer.}@*)</answer>
\end{lstlisting}
\end{methodbox}

\begin{reviewrevision}
We train the policy on the supervised dataset
$\mathcal{D}_{\mathrm{SFT}}$ of image--instruction--target
response triples $(x,q,y^*)$, jointly supervising
the description, analysis, and answer. For target token
$y_t^*$ with prefix $y_{<t}^*$, the binary mask $m_t$ selects
completion tokens, assigning zero weight to the input, padding, and
fixed assistant prefix. We minimize the EoT-SFT loss $\mathcal L_{\mathrm{SFT}}(\theta)$,
the masked, token-averaged negative log-likelihood given by
\end{reviewrevision}

\begin{equation}
\begin{aligned}
\mathcal L_{\mathrm{SFT}}(\theta)
={}&-\mathbb E_{(x,q,y^*)\sim\mathcal D_{\mathrm{SFT}}}
\\
&\left[
\frac{\sum_t m_t\log\pi_\theta(y_t^*\mid x,q,y_{<t}^*)}
{\sum_t m_t}
\right].
\end{aligned}
\label{eq:eot-loss}
\end{equation}

\begin{reviewrevision}
The resulting policy $\pi_{\mathrm{S1}}$, also denoted $M_1$,
initializes RTPO and serves as its frozen KL reference.
\end{reviewrevision}

\subsection{Anchor--Bridging Supervision}
\label{sec:anchor-bridge}

\begin{reviewrevision}
Remote sensing scenes may share similar prominent objects while differing in local configurations that are decisive for task-specific judgments.
Stage~2 refines the Stage~1 process with task-matched scenes for which
the same question and conditions apply to each image on its own.
Changes in local attributes, connections, group membership, or visible
boundaries may alter supported judgments, while task-irrelevant
differences leave the task-specific judgment unchanged.

During pair construction, the Anchor image $x_A$ establishes a scene
configuration, and the Bridging image $x_B$ provides a comparable
configuration with a meaningful contrast in the evidence. Both sides
receive symmetric supervision. The shared instruction $q$
specifies the question, target, scope, and public conditions for both
images. Pair review checks task applicability, the relevant evidence
on each side, and the judgment relation that evidence supports.

Each image $x_i$, $i\in\{A,B\}$, has an Evidence Card
$\mathcal E_i$, and the pair shares a Boundary Card
$\mathcal B_{AB}$. Together, these components form the training unit
\end{reviewrevision}

\begin{equation}
\mathcal P_{AB}
=
(x_A,x_B,q,\mathcal E_A,\mathcal E_B,\mathcal B_{AB}).
\label{eq:pair-unit}
\end{equation}

\begin{reviewrevision}
\paragraph{Evidence Cards.}
Each $\mathcal E_i$ records visible objects, local attributes, spatial
relations, coverage priorities, and visibility limits, associating
observations with their objects or regions. Gravity uses the image
and its card to assess visual fidelity and description coverage.

\paragraph{Boundary Cards.}
The shared $\mathcal B_{AB}$ links observations to the task through
public conditions, supported inferences, inference limits, side-specific
answer contracts, and comparison annotations. Flight checks reasoning
from the generated descriptions and analyses using the public task,
with permitted comparison metadata for paired evaluation, while
Task Check evaluates the final answer using the contract assigned
to its image.

The set contains 500 pairs of distinct CPRS harbor images~\cite{fba},
with 1,000 images, 1,000 Evidence Cards, and 500 Boundary Cards.
The pairs include 372 decisive-change cases, 77 invariance cases,
and 51 visibility-scope cases. They cover changed judgments, consistency
supported by each image's own evidence, and narrower claims when the
task depends on obscured details.
Tasks cover local attributes, spatial layout, connections, counting,
functional interpretation, evidence sufficiency, comparison, and
visible activity.

Construction details appear in \cref{app:data,app:cards}.
\end{reviewrevision}

\subsection{Gravity and Flight Verification}
\label{sec:gravity}

Given a training unit $\mathcal P_{AB}$, the policy generates
response $y_i$ from image $x_i$ and the shared instruction $q$.
The joint response distribution is
\begin{equation}
p_\theta(y_A,y_B\mid x_A,x_B,q)
=\prod_{i\in\{A,B\}}\pi_\theta(y_i\mid x_i,q).
\label{eq:joint-policy}
\end{equation}
Verification separates visual reliability from conditional reasoning
(\cref{fig:framework}). Both verifiers use the same fixed
DeepSeek-V4.1-Flash judge through separate calls with mechanism-specific prompts. Each call receives only the permitted information, although
judging errors may remain correlated through the shared model.
The judge returns integer scores and brief justifications using the
criteria summarized in the boxes. View-specific anchors appear in
\cref{app:rubrics}, and the verifier implementation in
\cref{app:verifier-implementation}.

\paragraph{Gravity.}
Gravity checks visual fidelity throughout the response and
coverage only in the initial description, separating the accuracy
of visual claims from which scene content is addressed.
Visual fidelity uses the image as the reference and accepts clearly
visible facts beyond the card's entries. The card guides coverage
priorities.
An unresolved material conflict between the image and card makes
the affected judgment unverifiable.

We index G1 and G2 by
$\mathcal D_G=\{\mathrm{fidelity},\mathrm{coverage}\}$,
respectively.
The projection $\mathsf P_d(y_i)$ selects visual claims
throughout the response for fidelity and the initial
description for coverage.
The rubric verifier $\mathsf V_d^G$ returns an integer in
$\{0,1,2,3\}$ for each view against $(x_i,\mathcal E_i)$.
Nonnegative view weights $w_d^G$ sum to one.
The normalized Gravity score $G_i\in[0,1]$ for response
$y_i$ is defined as
\begin{equation}
G_i
=\sum_{d\in\mathcal D_G}w_d^G\,
\frac{\mathsf V_d^G\!\left(x_i,\mathcal E_i;
             \mathsf P_d(y_i)\right)}{3}.
\label{eq:gravity}
\end{equation}

\begin{methodbox}{Gravity Rubric}
\begin{lstlisting}[basicstyle=\ttfamily\small]
REFERENCE image x and Evidence Card E
CANDIDATE initial description and all visual claims in the response

(*@\mbhead{G1 FIDELITY}@*)
  Use the image to check entities, attributes, counts, ownership, and
  spatial relations. Match certainty to visibility. Accept clear
  observations beyond the card.

(*@\mbhead{G2 COVERAGE}@*)
  Use the card's priorities to assess coverage of subjects, regions, attributes, and
  relations in the initial description only.

(*@\mbemph{SCORE G1 and G2 per response (0..3).}@*)
  (*@\mbemph{RETURN scores and evidence-based reasons.}@*)
\end{lstlisting}
\end{methodbox}

\paragraph{Flight.}
Flight evaluates reasoning conditional on the described visual facts.
Gravity checks those facts against the image.
F1 assesses evidence use within each analysis, while F2
examines task execution on both sides and the relationship
between their judgments.

For Flight, we use $u_i$ to represent the description
and analysis of response $y_i$. We index F1 and F2 by
$\mathcal D_F=\{\mathrm{use},\mathrm{adaptation}\}$,
respectively.
The corresponding candidate inputs are
$\mathcal Z_i^{\mathrm{use}}=u_i$ and
$\mathcal Z_i^{\mathrm{adaptation}}=(u_A,u_B)$.
The projection $\mathsf\Pi_d(\mathcal B_{AB})$ supplies the
public task conditions stated in $q$, with permitted comparison
applicability and type added for F2. The comparison type identifies
the intended check, while the described premises under the public
task determine the supported relation. Flight's complete input
comprises the selected candidate content, the public task, and the permitted
comparison fields. Images and Evidence Cards support Gravity, while
answer contracts and reference judgments support Task Check.
The verifier $\mathsf V_d^F$ returns an integer in
$\{0,1,2,3,4,5\}$.
Applicability follows the task and comparison annotations.
Nonnegative weights $w_d^F$ are normalized over applicable
views and set to zero otherwise. The normalized Flight score
$F_i\in[0,1]$ for side $i$ is defined as
\begin{equation}
F_i
=\sum_{d\in\mathcal D_F}w_d^F\,
\frac{\mathsf V_d^F\!\left(q,\mathsf\Pi_d(\mathcal B_{AB});
                  \mathcal Z_i^d\right)}{5}.
\label{eq:flight}
\end{equation}
F1 is scored per response, while F2 is scored jointly
for the pair, with a decisive contextual failure on either
side limiting the score.
Flight accepts reasoning that is valid conditional on a described
premise, including a premise that Gravity identifies as visually false.
A visually false claim introduced only in analysis can fail
Gravity for inaccuracy. If the analysis relies on that claim to
reach its conclusion, Flight can also penalize the missing
descriptive support.

\begin{methodbox}{Flight Rubric}
\begin{lstlisting}[basicstyle=\ttfamily\small]
REFERENCE public task and side conditions
          F2 adds relation type and applicability
PREMISES  take each description as given

(*@\mbhead{F1 EVIDENCE USE}@*)
  Trace decisive inferences to facts in the description. Check task rule and scope.

(*@\mbhead{F2 PAIRED ADAPTATION}@*)
  Check each side's target, rule, and scope. If the type label conflicts with the
  descriptions, follow the described evidence. Accept agreement when both support
  the same judgment and a change when a decisive fact differs. If that fact is
  unresolved, expect a less specific judgment on the affected side. A decisive
  failure on either side limits F2.

(*@\mbemph{SCORE F1 per side, F2 per pair (0..5).}@*)
  (*@\mbemph{RETURN scores and brief justifications.}@*)
\end{lstlisting}
\end{methodbox}

\subsection{Relational Transfer Policy Optimization}
\label{sec:rtpo}
\label{sec:rtpo-train}

RTPO updates the single-image policy initialized by $M_1$ using
relational feedback from the A--B units in \cref{eq:pair-unit}.

\paragraph{Sampling and Matching.}
A training unit contains an annotated image pair under a shared task,
within which sampled responses are matched into candidate pairs.
For each unit $\mathcal P_{AB}$, the rollout policy
$\pi_{\theta_{\mathrm{old}}}$ independently samples $K$ responses
from each image $x_i$ and the shared instruction $q$.
Random one-to-one matching uses each response exactly once to form
$K$ candidate pairs, with $y_{i,k}$ denoting side $i\in\{A,B\}$ of
pair $k\in\{1,\ldots,K\}$.
G1, G2, and F1 are scored per response, while F2 is scored
once per matched pair and shared by both sides.
The resulting $G_{i,k}$ and $F_{i,k}$ follow
\cref{eq:gravity,eq:flight}.

\paragraph{Pair Reward.}
Task Check evaluates each final answer against its side-specific
contract in $\mathcal B_{AB}$, giving normalized correctness
$C_{i,k}\in[0,1]$.
The binary indicator $v_{i,k}$ records whether the response contains
text in all three phases, ordered as description, analysis, and answer.
The task-correctness contribution is
$r_k^{\mathrm{task}}=
\frac{\lambda_C}{2}\sum_{i\in\{A,B\}}v_{i,k}C_{i,k}$,
with fixed $\lambda_C=0.50$.
A protocol violation removes that side's reward contribution while
preserving the valid contribution from the other side.

\begin{samepage}
With reward weights $\lambda_G$ and $\lambda_F$ for Gravity
and Flight, the total reward $R_k(G,F)$ for candidate pair
$k$ is defined as
\begin{equation}
R_k(G,F)=r_k^{\mathrm{task}}+
\frac12\sum_{i\in\{A,B\}} v_{i,k}
[\lambda_G G_{i,k}+\lambda_F F_{i,k}].
\label{eq:pair-reward}
\end{equation}
\end{samepage}

\paragraph{Shared Advantage.}
Within each A--B unit, let $\mathcal V$ index candidate pairs
with valid verifier outputs, whose rewards have mean
$\mu_{\mathcal V}$ and standard deviation $\sigma_{\mathcal V}$.
For each $k\in\mathcal V$ and stabilizing constant $\epsilon_A>0$,
the advantage is
\begin{equation}
\hat A_k(G,F)
=\frac{R_k(G,F)-\mu_{\mathcal V}}{\sigma_{\mathcal V}+\epsilon_A}.
\label{eq:shared-advantage}
\end{equation}
Both responses of pair $k$ receive the same group-normalized
advantage. Flight adaptation supplies the cross-scene assessment
in the pair reward.

\paragraph{Policy Update.}
The update follows GRPO~\cite{deepseekmath} with a clipped
policy loss~\cite{ppo} and importance ratios against
$\pi_{\theta_{\mathrm{old}}}$.
For each valid pair $k$, $\mathcal L_{i,k}(\hat A_k(G,F))$
denotes the clipped policy loss for response $y_{i,k}$ using
the shared advantage. It sums losses over the $T_{i,k}$
retained completion tokens and divides by
$\bar T_i=K^{-1}\sum_{j\in\mathcal V}T_{i,j}$.
Consequently, $K^{-1}\sum_k\mathcal L_{i,k}$ gives one
retained-token mean on side $i$.
Unverifiable pairs contribute zero to both the policy and KL losses.
Let $\Omega$ denote sampling of training units, responses, and
their matching.
A coefficient $\beta$ controls KL regularization toward the
frozen Stage-1 policy $\pi_{\mathrm{S1}}$.
We minimize the RTPO objective
$\mathcal J_{\mathrm{RTPO}}(\theta)$ defined by
\begin{equation}
\begin{aligned}
\mathcal J_{\mathrm{RTPO}}(\theta)
=\;&\mathbb E_{\Omega}\!\left[
\frac{1}{2K}\sum_{k=1}^{K}\sum_{i\in\{A,B\}}
\mathcal L_{i,k}\!\big(\hat A_k(G,F)\big)
\right]\\
&+\beta\,\mathbb E_{\Omega}\!\left[
D_{\mathrm{KL}}(\pi_\theta\,\|\,\pi_{\mathrm{S1}})
\right].
\end{aligned}
\label{eq:rtpo-objective}
\end{equation}
The outer factor weights the two sides equally after token
averaging. The reference remains fixed while the policy is updated.
Loss details, the sampled KL estimator, and group-validity
handling appear in \cref{app:training}.

\section{Experiments}
\label{sec:experiments}

\subsection{Experimental Setup}
\label{sec:setup}

\begin{reviewrevision}
\paragraph{Data and evaluation.}
Harbor training uses \sftcount{} EoT demonstrations and 500 task-matched
pairs from CPRS~\cite{fba}. We evaluate on HarborEval and
OpenEval~\cite{fba}, together with single-image harbor subsets of
VRSBench~\cite{vrsbench}, RSVQA~\cite{rsvqa}, OmniEarth~\cite{omniearth}, and
VLRS-Bench~\cite{vlrsbench}. Comparisons use the same fixed evaluation
items, and each benchmark retains its scoring protocol. Overall
averages the six scores on a 0--100 scale, assigning equal weight
to each benchmark. FBA specifies blinded expert review for OpenEval and
a fixed image-grounded judge for HarborEval reports~\cite{fba,mtbench}.
OpenEval ratings cover grounding, object coverage, functional zones,
spatial relations, uncertainty handling, and concision.
Data and evaluation details appear in
\cref{app:evaluation}.

\paragraph{Models and comparisons.}
We adapt Qwen3.5-4B~\cite{qwen35} with LoRA~\cite{lora}, freezing the
vision encoder and sampling $K=6$ candidates per image for RTPO.
All harbor RL ablations and matched controls start from the same
supervised checkpoint $M_1$,
with matched image--question exposure, generation limits, and
optimization settings. Removing a reward term leaves the other
coefficients fixed. All training variants are evaluated with
single-image inputs. Training and control details appear in
\cref{app:hyperparameters,app:matched-controls}.
\end{reviewrevision}

\subsection{\reviewrev{Comparison with Existing Models}}
\label{sec:main-results}
\label{sec:model-compare}

\reviewrev{HarborEval tests harbor expertise, while VRSBench and
RSVQA probe fine visual detail and relational question answering.
OpenEval assesses open-ended reporting, OmniEarth geospatial
understanding, and VLRS-Bench scene reasoning and decisions.
G\&F improves over its backbone on all six sets and leads five
(\cref{tab:model_comparison}). Its mean of 69.12 exceeds the backbone
by 9.76 points and RS-EoT, the strongest remote sensing baseline,
by 12.86. RSThinker leads on VRSBench (67.67), where G\&F makes its
largest gain over the backbone (+16.98).}

\begin{figure}[!tbp]
  \centering
  \includegraphics[width=\textwidth]{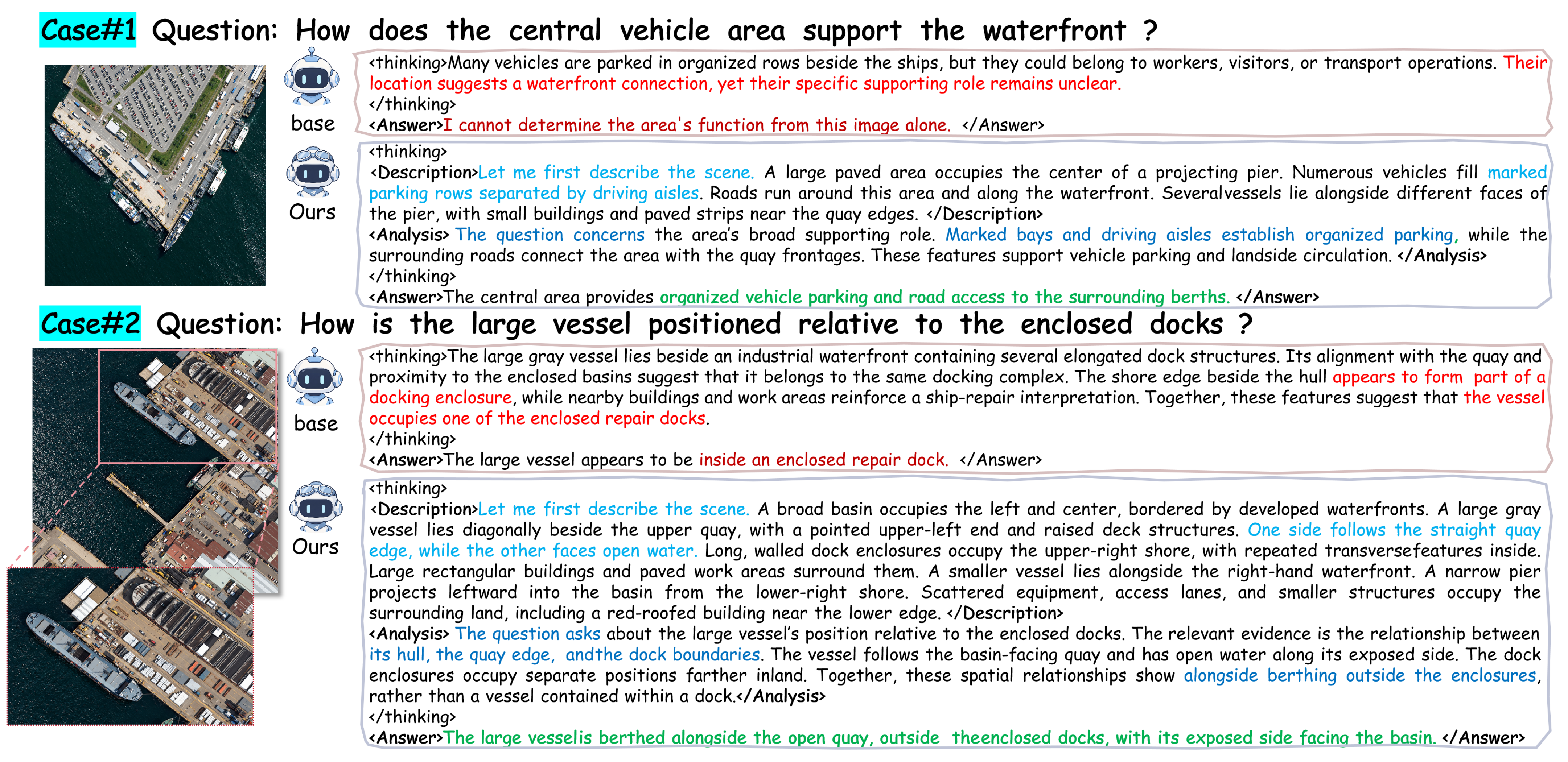}
  \caption{Qualitative comparison on held-out harbor images using the
  same image--question inputs. The cases illustrate functional inference
  and vessel--dock spatial reasoning, with relevant evidence highlighted.}
  \label{fig:qualitative-results}
\end{figure}

\reviewrev{\Cref{fig:qualitative-results} illustrates how the
responses connect visible observations to the requested judgments.
In Case~1, the base recognizes vehicle rows and leaves the area's
supporting role unresolved, whereas G\&F connects marked bays, driving aisles,
and quay access to organized parking and landside circulation,
answering the functional question at the level supported by the layout.
In Case~2, the base infers dock enclosure from nearby repair facilities.
G\&F instead uses the hull--quay alignment, exposed open water, and
separate dock boundaries to place the vessel alongside the open quay,
outside the enclosures.}

\begin{table}[t]
\centering
\caption{Comparison on six harbor evaluation sets. Overall is their
unweighted mean. Data Scale follows source-reported adaptation counts,
with n/r marking unreported values. G\&F uses \sftcount{} demonstrations and 500 pairs.
Bold and underlining indicate the best and second-best scores.}
\label{tab:model_comparison}

\setlength{\tabcolsep}{5.0pt}
\renewcommand{\arraystretch}{1.08}

\resizebox{\textwidth}{!}{
\begin{tabular}{lccccccccc}
\toprule

\textbf{Model}
&
\textbf{Params}
&
\textbf{Data Scale}
&
\textbf{Overall}
&
\textbf{HarborEval}
&
\textbf{VRSBench}
&
\textbf{RSVQA}
&
\textbf{OpenEval}
&
\textbf{OmniEarth}
&
\textbf{VLRS-Bench}
\\

\midrule

\multicolumn{10}{c}{
\textit{General Vision-Language Models}
}
\\
\midrule

LLaVA-1.5 [CVPR'24]~\citationrev{\cite{llava15}}
& 7B
& 1.2M
& \reviewrev{36.87}
& 46.22
& 37.44
& \reviewrev{27.26}
& \reviewrev{45.75}
& \reviewrev{32.32}
& \reviewrev{32.20}
\\

Qwen3.5 [Tech. Rep.'26]~\citationrev{\cite{qwen35}}
& 4B
& n/r
& \underline{59.36}
& 73.15
& 37.21
& \reviewrev{58.69}
& \underline{\reviewrev{67.34}}
& \underline{\reviewrev{80.39}}
& \reviewrev{39.39}
\\

\midrule

\multicolumn{10}{c}{
\textit{Remote Sensing VLMs}
}
\\
\midrule

GeoChat [CVPR'24]~\citationrev{\cite{geochat}}
& 7B
& 318K
& \reviewrev{41.61}
& 47.49
& 53.44
& 51.46
& 21.78
& \reviewrev{41.42}
& \reviewrev{34.09}
\\

\citationrev{SkySenseGPT [arXiv'24]~\cite{skysensegpt}}
& 7B
& 3.00M
& \reviewrev{39.25}
& 47.24
& 42.99
& 42.98
& 35.78
& \reviewrev{29.38}
& \reviewrev{37.12}
\\

EarthDial [CVPR'25]~\citationrev{\cite{earthdial}}
& 4B
& 11.11M
& \reviewrev{44.01}
& 65.02
& 42.33
& \reviewrev{53.96}
& \reviewrev{34.31}
& \reviewrev{38.50}
& \reviewrev{29.92}
\\

VHM [AAAI'25]~\citationrev{\cite{vhm}}
& 7B
& 1.58M
& \reviewrev{46.30}
& 57.62
& 49.70
& 52.13
& 54.44
& \reviewrev{32.44}
& \reviewrev{31.44}
\\

SkyEyeGPT [ISPRS'25]~\citationrev{\cite{skyeyegpt}}
& 7B
& 968K
& \reviewrev{25.98}
& 28.28
& 36.72
& 36.72
& 12.33
& \reviewrev{21.36}
& \reviewrev{20.45}
\\

LHRS-Bot-Nova [ISPRS'25]~\citationrev{\cite{lhrsbotnova}}
& 8B
& 2.02M
& \reviewrev{34.64}
& 39.73
& 28.40
& 31.15
& 22.96
& \reviewrev{48.48}
& \reviewrev{37.12}
\\

\midrule

\multicolumn{10}{c}{
\textit{Reasoning-Oriented RS VLMs}
}
\\
\midrule

RS-EoT [CVPR'26]~\citationrev{\cite{rseot}}
& 7B
& 65.7K
& \reviewrev{56.26}
& \underline{74.54}
& \reviewrev{31.86}
& \underline{\reviewrev{63.98}}
& \reviewrev{58.19}
& \reviewrev{67.34}
& \underline{\reviewrev{41.67}}
\\

RSThinker [ICLR'26]~\citationrev{\cite{rsthinker}}
& 9B
& 380K
& \reviewrev{55.02}
& 69.06
& \textbf{\reviewrev{67.67}}
& \reviewrev{50.07}
& \reviewrev{42.53}
& \reviewrev{66.33}
& \reviewrev{34.47}
\\

\midrule

\multicolumn{10}{c}{
\textit{Models trained with the proposed framework}
}
\\
\midrule

\oursrow
\textbf{G\&F(Qwen3.5) (Ours)}
& \textbf{4B}
& \textbf{2K}
& \textbf{69.12}
& \textbf{82.57}
& \underline{54.19}
& \textbf{67.18}
& \textbf{82.07}
& \textbf{80.99}
& \textbf{47.73}
\\

\bottomrule
\end{tabular}
}

\end{table}

\subsection{Component and Training Ablations}
\label{sec:analysis}

\begin{table}[!tb]
  \centering
  \caption{Harbor training comparisons. RL starts from $M_1$. Cost is the
measured training wall time, including Stage~1 and verifier waiting,
normalized to $M_1$ under the optimized implementation. Measurement
details in \cref{app:diagnostics}.}
  \label{tab:ablation}
  \begingroup\paperrevcolor
  \small
  \setlength{\tabcolsep}{0pt}
  \renewcommand{\arraystretch}{1.14}
  \begin{tabularx}{\columnwidth}{@{}>{\hspace*{3pt}\raggedright\arraybackslash}l
      >{\centering\arraybackslash}X
      >{\centering\arraybackslash}X>{\centering\arraybackslash}X@{}}
    \toprule
    \textbf{Setting} & \textbf{Overall} & \textbf{$\Delta$ vs $M_0$}
    & \textbf{Rel.\ cost ($\times$)} \\
    \midrule
    $M_0$ (base) & 59.36 & 0.00 & n/a \\
    $M_1$ (+EoT-SFT) & 64.56 & +5.20 & 1.00 \\
    \midrule
    \multicolumn{4}{c}{\textit{Paired reward ablations}} \\
    \midrule
    w/o G and F & 63.34 & +3.98 & 8.15 \\
    w/o Flight & 65.11 & +5.75 & 10.13 \\
    w/o Gravity & 66.32 & +6.96 & 15.83 \\
    \midrule
    \multicolumn{4}{c}{\textit{Matched training controls}} \\
    \midrule
    Single-image RL & 65.12 & +5.76 & 13.38 \\
    Paired RL w/o F2 & 66.79 & +7.43 & 13.57 \\
    Shuffled F2 feedback & 64.71 & +5.35 & 17.20 \\
    \joinrev{Joint-input RL} & \joinrev{66.49} & \joinrev{+7.13} & 16.86 \\
    \oursrule
    \oursrow \textbf{G\&F} & \textbf{69.12} & \textbf{+9.76} & \textbf{16.69} \\
    \bottomrule
  \end{tabularx}
  \endgroup
\end{table}

\begin{reviewrevision}
\Cref{tab:ablation} shows a 5.20-point gain from EoT-SFT and a
further 4.56 points from RTPO.

\paragraph{Reward components.}
\label{sec:flight-ablation}
Task-only RL scores 63.34, below $M_1$ (64.56). Removing Flight or
Gravity from full RTPO costs 4.01 or 2.80 points, respectively.

\paragraph{Flight-guided relational feedback.}
Paired RL without F2 combines local rewards, shares the advantage,
and checks validity jointly. It gains 1.67 points over single-image RL
at similar cost. Adding F2 within this setup yields another 2.33 points.

\paragraph{Relational feedback alignment.}
Shuffled F2 feedback retains image pairs and local scoring but
permutes F2 scores among each unit's six candidate pairs before
forming rewards and shared advantages. It reaches 64.71 at
17.20$\times$, 4.41 points below G\&F, 2.08 below Paired RL w/o F2,
and 0.15 above $M_1$, favoring aligned F2 feedback.

\paragraph{Generation context.}
Joint-input RL uses the same image pairs, verification criteria, and
reward weights as G\&F. Given both images, it generates a description,
analysis, and answer for each in a single sequence. The second response
can then draw on observations from the first as well as the visual
evidence in either scene. Under
single-image evaluation it reaches 66.49, 2.63 points below G\&F at
slightly higher training wall time (16.86$\times$ versus
16.69$\times$). The joint context may mix scene-specific evidence across
responses, weakening the supervision signal for fine-grained
distinctions. Control definitions appear in
\cref{app:matched-controls,app:joint-control}.

\end{reviewrevision}

\subsection{Token-Level Evidence Organization}
\label{sec:token-evidence-organization}

\begin{figure}[!t]
  \centering
  \includegraphics[width=0.9\linewidth]{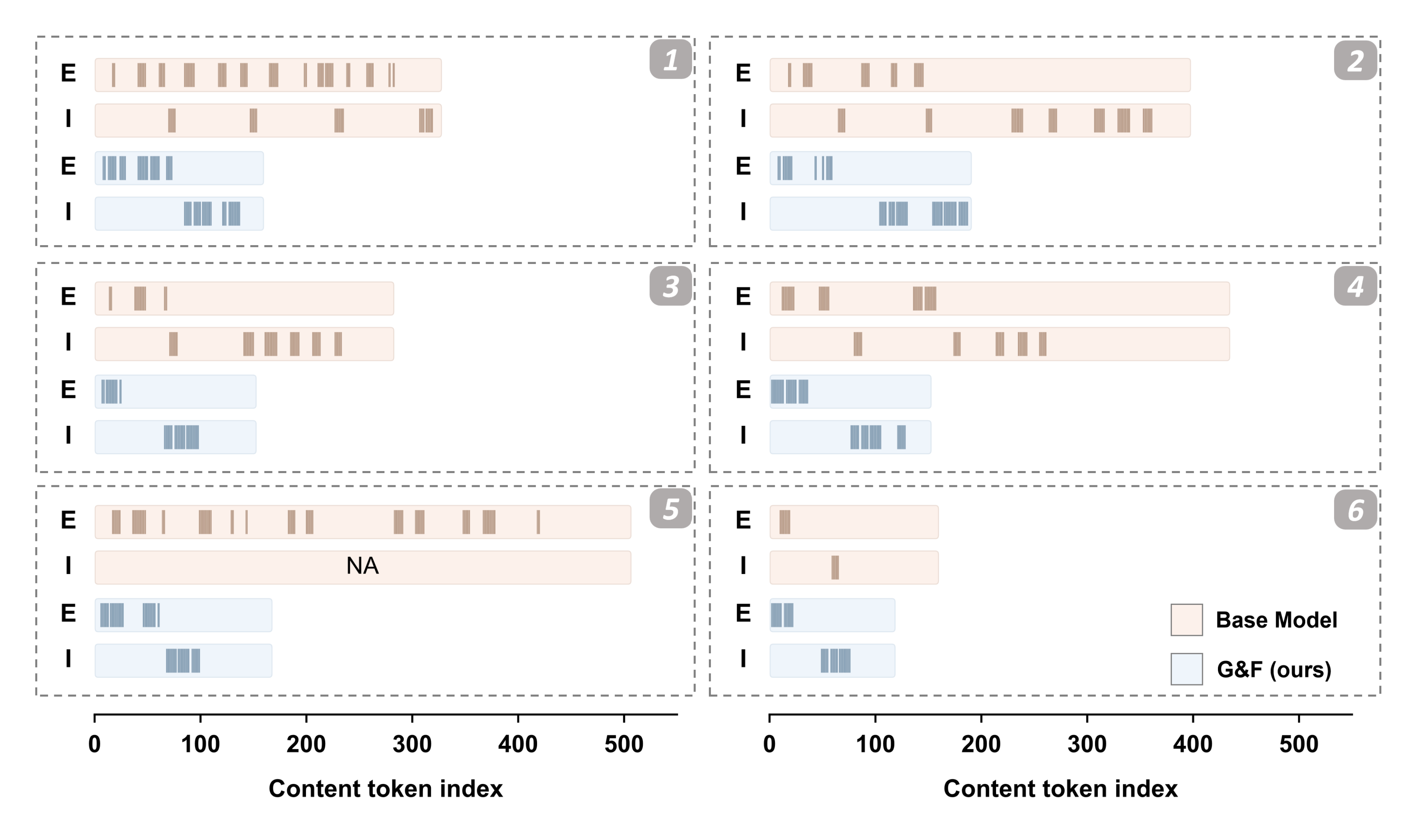}
  \caption{Token-level organization of six randomly sampled response
  pairs. Peach (Base) and blue (G\&F) tracks share a content-token axis
  that excludes format-only tokens. Dark marks show evidence (E) or
  inference (I) phrase membership, not attention. NA marks a response
  with no analytical tokens.}
  \label{fig:token-level-evidence}
\end{figure}

\Cref{fig:token-level-evidence} compares one Base and one G\&F
response per case on six randomly sampled image--question inputs,
each with two required visual facts and a reference judgment.
Evidence phrases and valid inference clauses are annotated with
parent sentences; identical sentences share labels across styles.
Densities exclude formatting tokens, include supported repetitions,
and weight cases equally.

G\&F states visible facts in the description and uses them in analysis.
The Base model distributes them across unstructured prose with partial
reuse in reasoning. Evidence density is 21.53\% for Base and 38.48\%
for G\&F. On the five cases with analytical spans in both responses,
inference density is 24.82\% and 46.61\%, respectively.
Mean task-related content density is 39.91\% for
G\&F and 22.35\% for Base, with mean full-response lengths of 179.50
and 355.50 tokens, respectively. Final answers agree with the reference
judgment in three of six Base responses and all six G\&F responses.
Required-fact coverage, counting each task's visual facts once,
is 91.67\% for Base and 100.00\% for G\&F.
\Cref{app:token-analysis} details the annotation and metrics.

\subsection{\reviewrev{Evaluation on Urban Scenes}}
\label{sec:cross-scene}

\reviewrev{Urban specialization uses 320 EarthVL-GLOBAL images
($1024\times1024$) from EarthVLSet~\cite{wang2026earthvl}, with
160 demonstrations, 60 training pairs, and 20 held-out development pairs.
A separate urban model undergoes Stage~1 and RTPO from the same base
checkpoint, retaining the harbor rubric dimensions and reward weights.
Evaluation uses fixed urban subsets of EarthVQA~\cite{earthvqa},
VRSBench~\cite{vrsbench}, and RSVQA HR~\cite{rsvqa};
data preparation and protocols are in \cref{app:urban}.}

\reviewrev{\Cref{tab:cross-scene} shows that +EoT-SFT improves all three
subsets over the base; +RTPO adds 1.93, 2.08, and 4.67 points,
respectively, supporting reuse beyond harbors.}

\begin{table}[htbp]
  \centering
  \caption{Urban specialization on fixed evaluation subsets.}
  \label{tab:cross-scene}
  \small
  \setlength{\tabcolsep}{0pt}
  \renewcommand{\arraystretch}{1.14}
  \begin{tabularx}{\columnwidth}{@{}>{\hspace*{3pt}\raggedright\arraybackslash}l
      >{\centering\arraybackslash}X
      >{\centering\arraybackslash}X>{\centering\arraybackslash}X@{}}
    \toprule
    \textbf{Stage} & \shortstack{\textbf{EarthVQA}\\\textbf{Urban}}
      & \shortstack{\textbf{VRSBench}\\\textbf{Urban}}
      & \shortstack{\textbf{RSVQA}\\\textbf{HR}} \\
    \midrule
    Base & 11.00 & 23.33 & 15.00 \\
    \rowcolor{stageurbanA}
    +EoT-SFT & 29.50 & 54.17 & 39.00 \\
    \rowcolor{stageurbanB}
    +RTPO & 31.43 & 56.25 & 43.67 \\
    \bottomrule
  \end{tabularx}
\end{table}

\FloatBarrier
\section{Conclusion}
\label{sec:conclusion}

\reviewrev{We presented Gravity and Flight for fine-grained remote
sensing specialization with limited supervision. EoT-SFT initializes
observation and analysis, and RTPO refines the policy on task-matched
pairs using Gravity's visual checks, Flight's conditional reasoning
checks, and task correctness. A shared pair-level advantage updates
the single-image policy. Harbor experiments show gains beyond
supervised initialization and favor aligned relational feedback.
Separate urban training also yields gains on all three subsets with
the same rubric dimensions and reward weights, supporting framework reuse.}

\section*{Declaration of competing interest}
The authors declare that they have no known competing financial
interests or personal relationships that could have appeared to
influence the work reported in this paper.

\section*{Acknowledgments}
This work is supported by the National Natural Science Foundation
of China under grants \mbox{62571271} and \mbox{62201552}, the
Natural Science Foundation of Tianjin under Grant No.\ \mbox{24JCQNJC01890},
and the Fundamental Research Funds for the Central Universities.

{\small
\bibliographystyle{unsrtnat}
\bibliography{main}
}
\clearpage
\section*{Supplementary Material}
\phantomsection
\addcontentsline{toc}{section}{Supplementary Material}
\appendix
\setcounter{table}{0}
\setcounter{figure}{0}
\renewcommand{\thetable}{S\arabic{table}}
\renewcommand{\thefigure}{S\arabic{figure}}
\setcounter{equation}{0}
\renewcommand{\theequation}{S\arabic{equation}}
\renewcommand{\theHequation}{supp.\arabic{equation}}

This supplement details the supervision data, verifier inputs and
scoring criteria, policy optimization, and evaluation protocols using
the notation of the main paper. A training unit is an annotated
Anchor--Bridging image pair, and a candidate pair consists of two
responses sampled and matched within that unit.

\section{Data Construction and Supervision}
\label{app:data}

\subsection{Evidence-of-Thought Demonstrations}
The harbor corpus contains \sftcount{} demonstrations over 888 images,
carrying 7,745 observation annotations, 1,411 uncertainty annotations,
and 1,698 alternative-interpretation assessments. Curation proceeds
through the three steps below to produce the description, analysis, and
answer phases of \cref{eq:eot-format}. Stage~2 adds 500 pairs over 1,000
images, each side with an Evidence Card and each pair with a Boundary
Card.

Scene observation first establishes an image-level record of principal
subjects, salient secondary objects, local attributes, counts, spatial
relations, and visibility limits. Observations are associated with
their objects or regions and include textual region references,
visibility, confidence, and coverage priorities. This record supplies
the visual premises from which task-specific targets are constructed.

Task interpretation then identifies the target, spatial scope, and public
conditions of the question. It selects relevant observations, applies
the requested operation, and states a judgment at the supported level
of specificity. Alternative interpretations are examined using the
details that distinguish them. Questions about the same image share
their initial description, while their analyses and answers reflect
different evidence selections and task conditions.

Finally, consistency review checks visual support for the observations, the
connection between the description and analysis, and agreement of the
answer with the question and the preceding reasoning. Uncertainty is
retained when a necessary detail is obscured or unresolved, while
broader conclusions are accepted when supported by the remaining
visible evidence.

\subsection{Anchor--Bridging Construction}
The harbor pairs use CPRS imagery introduced by FBA~\cite{fba}.
Each unit contains two distinct images to which the same question and
public conditions apply separately, and both sides receive symmetric
supervision. Pair review checks that the task is applicable to both
images and identifies which differences matter for the requested
judgment. Task coverage includes local attributes, layout, connections,
counting and grouping, functional interpretation, evidence sufficiency,
candidate comparison, and visible activity, with the largest groups
being objects and attributes (143 pairs) and spatial layout (131).
The 500 units partition by evidence relation in \cref{tab:pair-relations}.

\begin{table}[!tb]
  \centering
  \caption{Composition of the 500 harbor training pairs. Relations
  specify task-relevant comparisons, allowing varied wording for
  semantically equivalent judgments.}
  \label{tab:pair-relations}
  \small
  \setlength{\tabcolsep}{5pt}
  \begin{tabular}{@{}lrr@{}}
    \toprule
    \textbf{Relation} & \textbf{Pairs} & \textbf{Share} \\
    \midrule
    Decisive change & 372 & 74.4\% \\
    Invariance & 77 & 15.4\% \\
    Visibility scope & 51 & 10.2\% \\
    \midrule
    Total & 500 & 100.0\% \\
    \bottomrule
  \end{tabular}
\end{table}

\subsection{Evidence Relations}

A decisive-change unit alters evidence that determines the supported
judgment. An invariance unit changes other scene content while
preserving the relevant support. Each response establishes
its judgment from its own image. A visibility-scope unit changes the
availability of evidence needed for a claim, so the supported answer
may become less specific. Missing evidence is distinguished from
observed absence. These relations concern the semantic judgment
supported by the task, allowing varied wording for equivalent answers.

\subsection{Identity Screening and Review Records}
The preparation bundle contains 1,000 original image records.
File-content, decoded-RGB, and rotation/reflection checks confirm
unique identities within Stage~2 and distinct identities across
Stage~1 and Stage~2. Harbor admission requires identifiable port-related
support, such as berthing, cargo handling, dock or shipyard access, or
sheltered small-craft service facilities. Observations are checked
against the original image and then reviewed together with the task,
cards, and evidence contrast; corrections address task-relevant
distinctions such as complete-hull counting, crossing structures versus
channels, object ownership, and crop-dependent scope. The records
document model-assisted annotation review, and all 500 units are
assigned to training, with evaluation supplied by the benchmarks of
\cref{app:evaluation}.

\subsection{Data Splits and Leakage Audit}
\label{app:data-splits}

For all six harbor benchmarks, including the final HarborEval split,
the evaluation source images are disjoint from those used in
Stage~1 demonstrations or on either side of a Stage~2 training pair.

For urban specialization, adaptation and evaluation use disjoint
source images, with related views assigned to the same partition
(\cref{app:urban}). Geographic overlap and exposure during backbone
pretraining are outside the scope of this audit.

\section{Card Contents and Verifier Views}
\label{app:cards}

\paragraph{Annotation records.}
An Evidence Card $\mathcal E_i$ organizes observations about image
$x_i$, with entity and region associations preserving which object owns
an attribute, participates in a relation, or contributes to a count.
The shared Boundary Card $\mathcal B_{AB}$ supplies the public task and
conditions, side-specific evidence licenses and inference limits,
answer contracts, and comparison annotations; the schema below lists the
fields of both. Each verification view selects only its permitted
fields, and all conditions needed to solve the task appear in the public
instruction.

\begin{methodbox}{Annotation Card Contents}
\begin{lstlisting}[basicstyle=\ttfamily\footnotesize]
EvidenceCard[i]:
  observations:
    entities, attributes, counts, relations
    object / region associations
    textual region references
    visibility, confidence, coverage priority
  coverage:
    main subjects, salient details, relations
  uncertainty:
    occlusion, crop limits, unresolved identity

BoundaryCard[AB]:
  task: shared question and public conditions
  sides[i]:
    evidence_licenses
    inference_boundaries
    answer_contract
  context_comparison:
    applicability, change_type
    paired expectations and supported scope
\end{lstlisting}
\end{methodbox}

\paragraph{Input separation.}
The policy receives only its own image $x_i$ and the shared instruction
$q$; cards and comparison annotations enter verification only after
response generation. Flight's candidate inputs are the description and
analysis $u_i$, with the public task and, for F2, comparison metadata as
reference, while Task Check scores the extracted answer against its
side-specific contract. \Cref{tab:verifier-inputs} lists the inputs of
each view.

\begin{table}[!t]
  \centering
  \caption{Information supplied to generation and verification.
  Public conditions are included in the policy instruction.
  Each scoring call receives its own prescribed view.}
  \label{tab:verifier-inputs}
  \small
  \setlength{\tabcolsep}{4pt}
  \renewcommand{\arraystretch}{1.12}
  \begin{tabular}{@{}p{0.13\textwidth}p{0.32\textwidth}p{0.48\textwidth}@{}}
    \toprule
    \textbf{View} & \textbf{Generated or assessed content} & \textbf{Available reference inputs} \\
    \midrule
    Policy & Response generated independently for one image &
      Own image and shared question with public conditions \\
    Gravity G1 & Visual assertions across description, analysis, and answer &
      Own image and Evidence Card \\
    Gravity G2 & Initial description only &
      Own image and the card's coverage anchors \\
    Flight F1 & Own description and analysis &
      Shared question and public task conditions \\
    Flight F2 & Both descriptions and analyses &
      Shared question, public conditions, comparison applicability and type \\
    Task Check & Extracted final answer &
      Question and side-specific answer contract \\
    \bottomrule
  \end{tabular}
\end{table}

\paragraph{Comparison metadata and conditional premises.}
The F2 metadata specifies comparison applicability and a
\texttt{change\_type} of decisive change, invariance, or visibility scope.
The type identifies the intended comparison, while the descriptions
supply its evidential premises. When the type conflicts with the
relation supported by those premises under the public task, the
described premises take priority.
Private paired expectations, evidence licenses, inference boundaries,
and answer contracts remain outside the Flight view.

This separation permits distinct diagnoses. A false fact already
established in the description is a Gravity error; an inference can
still be valid conditional on that fact. For Flight, decisive visual
details used in analysis require support in the preceding description,
including details that are true of the image. An image-false claim introduced only
in analysis and used decisively can fail both visual fidelity and
evidence use.

\paragraph{Image support and card coverage.}
The image is the reference for visual truth, and the Evidence Card
anchors observations and coverage priorities. Fidelity accepts a
clearly visible fact even when it falls outside the card's entries.
A material image--card conflict that remains unresolved from the
supplied inputs makes the affected judgment unverifiable. Coverage
follows the card's priorities, while fidelity checks the response's
visual assertions against the image.

\subsection{Verifier Implementation}
\label{app:verifier-implementation}

Gravity and Flight use the same fixed DeepSeek-V4.1-Flash judge,
as specified in \cref{sec:gravity}, through separate calls with
view-specific criteria. Each call receives its prescribed inputs,
with scores and explanations kept separate across calls. The views
differ in information access and rubric; judging errors may remain
correlated through the shared model.

Response phases are selected deterministically from the structured output. Gravity
receives the response sections needed to inspect visual assertions
directly. Flight receives only the selected description and analysis
fields. The projections $\mathsf P_d$ and $\mathsf\Pi_d$ in the main
paper specify candidate content and permitted reference fields,
respectively. Brief reasons accompany integer scores and connect the
assessment to supplied observations, public conditions, or missing
support. Candidate response text is supplied as evaluation data.

Scores are validated against the scale of their dimension before
normalization. Gravity scores are divided by three, Flight scores by
five, and task correctness uses its own normalization. A raw score
therefore has meaning only within its corresponding rubric.
Protocol validity and an unverifiable scorer output are handled
separately, as described in \cref{app:rubrics,app:training}.

\section{Rubric Criteria and Score Anchors}
\label{app:rubrics}

\subsection{Scoring Criteria and Validity}
\label{app:scoring-criteria}

The following anchors summarize the scoring criteria used in the
method. G1 and G2 use integers in $\{0,1,2,3\}$; F1 and F2 use
integers in $\{0,1,2,3,4,5\}$. Scores reflect how a visual or reasoning
defect affects the task. The anchors apply across response lengths,
wording choices, and numbers of reasoning steps.

\paragraph{G1 Visual fidelity.}
A score of \textbf{3} denotes accurate visual claims with certainty
appropriate to visibility. \textbf{2} denotes a minor local inaccuracy
while the principal scene facts remain reliable. \textbf{1} denotes a
material error in an entity, attribute, count, ownership, extent, or
spatial relation. \textbf{0} denotes predominantly incorrect or
fabricated scene facts. The inspected claims may occur in any response
phase. Each claim requires visual support commensurate with its
specificity and certainty.

\paragraph{G2 Description coverage.}
A score of \textbf{3} denotes coverage of the main subjects, salient
local details, and key relations identified by the Evidence Card.
\textbf{2} denotes useful coverage with an important omission.
\textbf{1} denotes a narrow account missing much of the scene structure
or salient local evidence. \textbf{0} denotes an empty or entirely
nonsubstantive visual description. Coverage records which priority
aspects are addressed, while G1 evaluates their accuracy. Concise
descriptions can satisfy the coverage priorities using their own wording.

\paragraph{F1 Evidence use.}
Equivalent wording, cross-sentence links, grouping, arithmetic, and
ordinary supported inferences are accepted, and missing support is
distinguished from a definite contradiction or an incorrect task
operation. A decisive visual claim used in analysis requires support in
the preceding description. \Cref{tab:flight-anchors} gives the six
levels.

\paragraph{F2 Paired adaptation and task execution.}
Both equal and different conclusions are valid when their described
evidence supports them, as are evidence-limited judgments when a
necessary detail is unavailable. One F2 score is assigned to each
candidate pair, and a decisive failure on either side limits it even
when the other side executes its task correctly.

\begin{table}[!t]
  \centering
  \caption{Summary of Flight score anchors. Both dimensions accept
  described observations conditionally. A missing decisive premise
  differs from an explicitly wrong task operation.}
  \label{tab:flight-anchors}
  \small
  \setlength{\tabcolsep}{4pt}
  \renewcommand{\arraystretch}{1.14}
  \begin{tabular}{@{}cp{0.43\textwidth}p{0.43\textwidth}@{}}
    \toprule
    \textbf{Score} & \textbf{F1: within-response evidence use} &
      \textbf{F2: paired task execution and adaptation} \\
    \midrule
    5 & Sufficient prior evidence is correctly selected and applied;
      the supported judgment has no task-use defect. &
      Both sides execute their applicable conditions, and the
      relation between judgments follows their described evidence. \\
    4 & All required links are supported, with only a minor,
      nondecisive overstatement in the reasoning. &
      Both sides satisfy the decisive and requested conditions,
      with only a minor supporting overstatement. \\
    3 & The decisive chain is supported, but a requested secondary
      aspect is unfinished or an ancillary operation is wrong. &
      Decisive conditions are met, but a secondary contextual
      requirement is incomplete. \\
    2 & A necessary prior observation, primary scope, or decisive
      operation is missing, including support first supplied in analysis. &
      A decisive applicable condition or its descriptive witness
      is missing on either side. \\
    1 & The analysis performs a definite wrong core operation,
      target selection, ownership assignment, or eligibility decision. &
      Either side violates a decisive public condition or uses an
      inapplicable target, rule, owner, or scope; or the judgment
      relation contradicts the described premises. \\
    0 & There is no usable task-directed analysis. &
      There is no usable paired task execution. \\
    \bottomrule
  \end{tabular}
\end{table}

\paragraph{Task correctness.}
The answer contract specifies acceptable semantics, required elements,
and task-specific score anchors. Exact-answer tasks use raw scores of
0 or 3; graded tasks use integers in $\{0,1,2,3\}$. Division by three
gives $C_{i,k}\in[0,1]$. Equivalent concise answers are accepted.
For label and counting tasks, Task Check scores the extracted answer;
when an explanation is requested, its necessary content forms part
of the contract. This component assesses the actual answer, whereas
Flight assesses reasoning in the description--analysis view.

\paragraph{Applicability and validity.}
Dimension applicability follows the public task and comparison
annotations. A required observation omitted from a response is scored
as a deficiency within the applicable dimension.
Inapplicable dimensions receive zero weight, and the remaining
weights are normalized within their verifier.
Controlled ablations set the removed term's coefficient to zero and
retain all other coefficients at their full-setting values.

Scoring distinguishes response quality, protocol compliance, and
assessability. A wrong inference or missing decisive premise can
receive a valid low integer score and remain in the reward statistics.

The response protocol requires nonempty description, analysis, and
answer phases in that order. Its binary indicator $v_{i,k}$ multiplies
the side's composed reward before pair averaging. A protocol violation
therefore removes that side's reward contribution while preserving
the other side's contribution when scoring remains verifiable.
A required assessment is unverifiable when the permitted inputs or
returned scoring record leave the score unresolved. Its
candidate pair is excluded from reward statistics and assigned zero
weight in both losses. Likelihood evaluation uses the original
generated tokens.

\subsection{Verifier Reliability Study}
\label{app:verifier-reliability}

\paragraph{Protocol.}
The audit draws 24 training units uniformly at random from the 500
pairs under a fixed seed and keeps one stored candidate pair per unit.
G1, G2, and F1 are scored on the 48 side responses, and F2 on the 24
candidate pairs. Two annotators score each item independently, using the
rubric anchors of \cref{app:rubrics} and only the inputs that the view
is permitted to see (\cref{tab:verifier-inputs}). They see neither the
judge scores nor the model identity. Disagreements are adjudicated to
one integer reference score per item. Each item also receives two judge
calls with identical inputs, prompts, and decoding settings.

\paragraph{Metrics.}
Agreement is the fraction of first-call judge scores within one point
of the adjudicated reference. Rates are reported separately on the
Gravity ($0$--$3$) and Flight ($0$--$5$) rubric scales. The repeat-change
rate counts any score difference between two calls with identical
inputs and settings, measuring scoring repeatability on fixed inputs.
Items whose required assessment is unscorable under the permitted inputs
are excluded from both rates and counted separately.

\begin{table}[!tb]
  \centering
  \caption{Verifier reliability on 24 training units. G1, G2, and F1
  cover 48 side responses; F2 covers 24 candidate pairs. Agreement
  counts a judge score as matching when it is within one point of the
  adjudicated reference, whereas the repeat-change rate counts any
  difference between two identical calls.}
  \label{tab:verifier-reliability}
  \small
  \setlength{\tabcolsep}{4pt}
  \begin{tabular}{@{}lccc@{}}
    \toprule
    Dimension & Items & \shortstack{Within one point\\of humans (\%)} &
      \shortstack{Repeat-change\\rate (\%)} \\
    \midrule
    G1 & 48 & 91.67 & 22.92 \\
    G2 & 48 & 100.00 & 12.50 \\
    F1 & 48 & 89.58 & 27.08 \\
    F2 & 24 & 83.33 & 29.17 \\
    \bottomrule
  \end{tabular}
\end{table}

\paragraph{Interpretation.}
Agreement describes correspondence with the adjudicated reference,
while repeat changes indicate variability for an unchanged response.
Such variability can perturb the relative rewards used to compute the
advantage. In this sample, agreement is highest for
description coverage, which is checked against explicit anchors inside a
single response, is lower for fidelity and evidence use
(89.58--91.67\%), and is lowest for paired adaptation (83.33\%), which
requires a relational judgment across two responses. The two rates order
the four dimensions consistently: the dimension with the highest
agreement also repeats most stably, and the lowest-agreement dimension
changes most often. Neither rate
validates the rubric itself, and 24 units support an indication of
reliability rather than a precise estimate.

\section{Optimization Details}
\label{app:training}

\paragraph{Supervised completion loss.}
EoT-SFT supervises the assistant completion containing the description,
analysis, and answer. The mask $m_t$ in \cref{eq:eot-loss} assigns
zero weight to input tokens, padding, and the fixed assistant prefix.
The completion-token sum is divided by the number of selected tokens
for that example. The resulting policy $M_1$ initializes RTPO and
provides its frozen KL reference.

\paragraph{Random matching and pair-level credit.}
The two images and their shared task are fixed before rollout. RTPO
samples $K=6$ responses per image and draws a random one-to-one match
between the two sets. Each response enters one pair with its own G1,
G2, F1, and Task Check scores. F2 is scored on the six realized pairs
out of the 36 possible combinations. Both members receive the same
advantage, so a response with full task credit can receive a negative
advantage when its pair reward falls below the unit mean.

\paragraph{Validity and normalization.}
Scorable errors enter the reward statistics. A format violation sets the
affected side's reward contribution to zero through $v_{i,k}$.
An unverifiable required assessment excludes the pair from normalization
and from both the policy and KL losses.

Let $\mathcal V$ contain pairs for which all applicable scoring
outputs are valid. Reward mean and standard deviation are computed
over this set within the training unit. For a retained group with
$m=|\mathcal V|\geq2$, the implementation uses population statistics,
\begin{equation}
\begin{aligned}
\mu_{\mathcal V}&=\frac1m\sum_{k\in\mathcal V}R_k,\\
\sigma_{\mathcal V}^{2}&=\frac1m\sum_{k\in\mathcal V}
(R_k-\mu_{\mathcal V})^2.
\end{aligned}
\label{eq:supp-group-statistics}
\end{equation}
Groups with fewer than two verifiable pairs or degenerate reward
variation are skipped. Each retained pair uses the scalar advantage
in \cref{eq:shared-advantage}.

\paragraph{Observed credit assignment.}
We examined an archived 100-unit harbor RTPO training run initialized
from $M_1$, using the recorded match indices and the advantages applied
in the updates. All 600 candidate pairs were retained, with no
unverifiable pairs or skipped units. Of the 200 image sides, 65, 27, 13,
12, 23, 17, and 43 had zero through six full-credit responses,
respectively. In a separate check on eight units before any RTPO update,
the frozen $M_1$ produced at least one full-credit response on 10 of 16
image sides.

\begin{table}[!tb]
  \centering
  \caption{Realized matches in the 100-unit training trace.
  A full-credit side has Task Check $C_{i,k}=1$. Reward and advantage
  are medians with the 25th and 75th percentiles in brackets.}
  \label{tab:observed-credit}
  \small
  \setlength{\tabcolsep}{2.4pt}
  \begin{tabular}{@{}crcc@{}}
    \toprule
    \shortstack{Full-credit\\sides} & \shortstack{Pairs\\(\%)} &
    Reward $R_k$ & Advantage $\hat A_k$ \\
    \midrule
    2 & 167 (27.8) & $0.90\,[0.85,0.94]$ & $0.66\,[0.22,1.09]$ \\
    1 & 190 (31.7) & $0.60\,[0.55,0.70]$ & $-0.15\,[-0.72,0.51]$ \\
    0 & 243 (40.5) & $0.53\,[0.41,0.61]$ & $-0.41\,[-1.10,0.53]$ \\
    \bottomrule
  \end{tabular}
\end{table}

Pairs with two full-credit responses had the highest median reward and
advantage (\cref{tab:observed-credit}). Among the 190 pairs with just
one full-credit response, 110 (57.9\%) received a negative shared
advantage. Estimating how often a different partner would reverse that
sign would require scoring alternative matches.

\paragraph{Clipped policy surrogate.}
At token $t$ of response $y_{i,k}$, the policy likelihood ratio is
\begin{equation}
\rho_{i,k,t}=
\frac{\pi_\theta(y_{i,k,t}\mid x_i,q,y_{i,k,<t})}
{\pi_{\theta_{\mathrm{old}}}(y_{i,k,t}\mid x_i,q,y_{i,k,<t})}.
\label{eq:supp-ratio}
\end{equation}
With clipping threshold $\epsilon_c>0$ and $\hat A_k=\hat A_k(G,F)$,
the token loss is the minimization-form clipped surrogate
$\ell_{i,k,t}=-\min\{\rho_{i,k,t}\hat A_k,
\operatorname{clip}(\rho_{i,k,t},1-\epsilon_c,1+\epsilon_c)\hat A_k\}$
applied to the retained tokens of \cref{eq:rtpo-objective}. Let
$T_{i,k}$ count those tokens and
$\bar T_i=K^{-1}\sum_{j\in\mathcal V}T_{i,j}$. The response mean then
reduces to the ratio of the summed token losses to the summed
retained-token counts over $\mathcal V$, so the nominal factor $1/K$
cancels and each retained token carries equal weight within its side.
The outer factor $1/(2K)$ in the main objective therefore averages
retained tokens per side and gives the two sides equal weight. RTPO
keeps the clipped update and group-relative baseline
of GRPO~\cite{deepseekmath}; its supervision is organized through
the pair reward and the shared advantage.

\paragraph{Sampled KL penalty.}
For each retained token, let
\begin{equation}
\begin{aligned}
\delta_{i,k,t}={}&\log\pi_{\mathrm{S1}}(y_{i,k,t}\mid x_i,q,y_{i,k,<t})\\
&-\log\pi_\theta(y_{i,k,t}\mid x_i,q,y_{i,k,<t}).
\end{aligned}
\label{eq:supp-log-ratio}
\end{equation}
The nonnegative sampled penalty is
$\exp(\delta_{i,k,t})-\delta_{i,k,t}-1$. It uses the same retained
token and candidate-pair masks as the policy loss, averages tokens
separately on each side, and then weights the sides equally.
The Stage-1 reference remains fixed throughout RTPO.

\subsection{Matched Single-Image and Paired Controls}
\label{app:matched-controls}

The controls share $M_1$, the image--question exposure, local rubric
criteria, task-correctness contracts, generation limits, and
optimization settings. Reward ablations set the removed mechanism's
coefficient to zero, leaving the remaining coefficients fixed.
Removing Flight removes both F1 and F2; the task-only setting removes
both Gravity and Flight while retaining task correctness.

Let $U_{i,k}$ be normalized F1 and $H_k$ normalized F2. With
$\gamma_U=\lambda_F w^F_{\mathrm{use}}$ and
$\gamma_H=\lambda_F w^F_{\mathrm{adaptation}}$, the local reward is
\begin{equation}
r^{\mathrm{loc}}_{i,k}
=v_{i,k}\left[0.50C_{i,k}+\lambda_G G_{i,k}
+\gamma_U U_{i,k}\right].
\label{eq:supp-local-reward}
\end{equation}
These coefficients follow the applicability weights of full RTPO.
The F1 coefficient stays fixed when paired adaptation is removed.

With both protocols valid, F2 is shared by both responses but counted
once after pair averaging, so a value constant across retained
candidates adds a common offset and cancels in the centered advantage;
only differences among candidates within a unit affect the update.
\Cref{tab:control-definitions} lists the terms used when both Flight
dimensions apply and the response protocol is valid.

\begin{table}[!tb]
  \centering
  \caption{Matched reward definitions. $C$, $G$, and $U$ are local
  scores; $H$ is shared paired adaptation. Coefficients are expressed
  using the notation of the main method. Shuffled F2 feedback permutes
  the association of F2 scores with candidate pairs.}
  \label{tab:control-definitions}
  \small
  \setlength{\tabcolsep}{3pt}
  \begin{tabular}{@{}lccccc@{}}
    \toprule
    \textbf{Setting} & $C$ & $G$ & $U$ & $H$ & \textbf{Advantage} \\
    \midrule
    Single-image RL & .50 & $\lambda_G$ & $\gamma_U$ & 0 & Per image \\
    Paired RL w/o F2 & .50 & $\lambda_G$ & $\gamma_U$ & 0 & Shared pair \\
    Shuffled F2 feedback & .50 & $\lambda_G$ & $\gamma_U$ & $\gamma_H$ & Shared pair \\
    Joint-input RL & .50 & $\lambda_G$ & $\gamma_U$ & $\gamma_H$ & Joint group \\
    G\&F (RTPO) & .50 & $\lambda_G$ & $\gamma_U$ & $\gamma_H$ & Shared pair \\
    \bottomrule
  \end{tabular}
\end{table}

Single-image RL computes group-relative advantages separately from
$\{r^{\mathrm{loc}}_{A,k}\}_{k=1}^K$ and
$\{r^{\mathrm{loc}}_{B,k}\}_{k=1}^K$. Both images remain in the
training exposure, and each response's reward and advantage depend
on its own image's candidates. Validity filtering and reward
normalization also operate independently, using separate retained
sets $\mathcal V_A$ and $\mathcal V_B$.

Paired RL w/o F2 keeps the shared pair-level advantage but sets
$\gamma_H=0$, so the two responses of a unit still share one advantage
while the paired-adaptation score is removed. Paired settings retain or
exclude candidate pairs jointly. The difference between Single-image
RL (65.12) and this paired formulation (66.79) includes the paired
reward, shared advantage, and joint validity handling. Full G\&F
(69.12) adds F2 within the paired formulation, with all applicable
scoring outputs entering its validity check.

Shuffled F2 feedback uses the constructed image--question inputs and
the rollout-matching procedure of full RTPO. Gravity, F1, and Task
Check scores retain their original response associations. F2 scores
are computed from the paired views, and their assignment among the six
candidate pairs within the same training unit is permuted before reward
composition and advantage calculation. The reward coefficients and optimization settings follow
full RTPO.

Joint-input RL retains the full verification criteria and coefficients
but generates both responses in one context. After generation,
side-specific fields are selected for the same scoring views.
\Cref{app:joint-control} details the generation-context comparison.
All settings use single-image inputs at evaluation.

\paragraph{Reading the controls together.}
The controls form a staged comparison. The independent local
formulation scores 65.12, the paired formulation without F2 scores
66.79, and full RTPO scores 69.12. The successive gains are 1.67 and
2.33 points under the corresponding reward and validity protocols.
The experiment with F2 uses the shared pair-level advantage.
Shuffled F2 feedback scores 4.41 points below full RTPO and 2.08 below
Paired RL w/o F2, while remaining 0.15 above $M_1$. Under this protocol,
misaligned relational feedback performs worse than omitting F2.
A joint generation context scores 2.63 points below full RTPO.

\subsection{Hyperparameters and Training Configuration}
\label{app:hyperparameters}

\Cref{tab:training-configuration} reports the recorded harbor training
settings and reward hyperparameters. Both stages adapt the language
layers of Qwen3.5-4B with LoRA while freezing the vision encoder.
The adapters cover self-attention projections
\texttt{q\_proj}, \texttt{k\_proj}, \texttt{v\_proj}, and
\texttt{o\_proj}; linear-attention projections
\texttt{in\_proj\_qkv}, \texttt{in\_proj\_z}, and \texttt{out\_proj};
and MLP projections \texttt{gate\_proj}, \texttt{up\_proj}, and
\texttt{down\_proj}. The LoRA rank, scaling parameter, and dropout
are 16, 32, and 0.05 in both stages.

\begin{table}[htbp]
  \centering
  \caption{Recorded harbor training configuration (top) and reward and
  optimization hyperparameters (bottom). The Stage-2 batch unit is an
  annotated image pair; each image contributes six rollouts.
  Within-view weights refer to the setting where both dimensions apply.}
  \label{tab:training-configuration}
  \small
  \setlength{\tabcolsep}{5pt}
  \renewcommand{\arraystretch}{1.08}
  \begin{tabular}{@{}lll@{}}
    \toprule
    Setting & EoT-SFT & RTPO \\
    \midrule
    Training data & \sftcount{} demonstrations & 500 image pairs \\
    Initialization & Base checkpoint & Stage-1 checkpoint \\
    Optimizer & Fused AdamW & AdamW \\
    Learning rate & $10^{-4}$ &  $2\times10^{-6}$ \\
    Schedule / warmup & Cosine / 5\% & Constant / 0 \\
    Training passes / updates & 3 epochs / 189 & 1 pass / 250 \\
    Batch per GPU & 1 demonstration & 1 image pair \\
    Gradient accumulation & 8 & 1 \\
    Effective global batch & 16 demonstrations & 2 image pairs \\
    Updates per rollout batch & Not applicable & 1 \\
    Weight decay / gradient norm cap & 0.01 / 1.0 & 0.01 / 1.0 \\
    Maximum sequence length & 4,096 tokens & 4,096 tokens \\
    Completion limit per response & Not applicable & 768 tokens \\
    Candidates per image & Not applicable & 6 \\
    Temperature / top-$p$ & Not applicable & 0.6 / 0.9 \\
    Precision / attention & BF16 / SDPA & BF16 / SDPA \\
    Gradient checkpointing & Enabled & Enabled \\
    Hardware & $2\times$ RTX A6000 & $2\times$ RTX A6000 \\
    Random seed & 42 & 42 \\
    \bottomrule
  \end{tabular}

  \vspace{0.9em}
  \begin{tabular}{@{}lc@{}}
    \toprule
    Hyperparameter & Value \\
    \midrule
    Task weight $\lambda_C$ & 0.50 \\
    Gravity weight $\lambda_G$ & 0.35 \\
    Flight weight $\lambda_F$ & 0.15 \\
    G1 / G2 weights & 0.75 / 0.25 \\
    F1 / F2 weights & 0.40 / 0.60 \\
    KL coefficient $\beta$ & 0.04 \\
    Advantage stabilizer $\epsilon_A$ & $10^{-6}$ \\
    Policy clipping threshold $\epsilon_c$ & 0.20 \\
    \bottomrule
  \end{tabular}
\end{table}

The two data columns correspond exactly to the supervision counts of
\cref{app:data}: three passes over the \sftcount{} demonstrations at an
effective batch of 16 give the recorded 189 updates, and one pass over
the 500 pairs at two pairs per update gives the recorded 250 updates.

These values make the pair reward of Eq.~(7) bounded by one: task
correctness contributes at most $0.5$, and the Gravity and Flight terms
contribute at most $0.5$ in total. Within each verifier the applicable
dimension weights are normalized to one, so $G_i$ and $F_i$ share a
common $[0,1]$ scale regardless of which dimensions apply. Removing a
term in the ablations sets its coefficient to zero and leaves the
remaining coefficients at the values reported here.

The recorded Stage-2 continuation processes the first 100 pairs at
$10^{-6}$ and the remaining 400 at $2\times10^{-6}$, for 250 global
updates in total. Each update processes two annotated pairs, each with
six candidate pairs, so one pass covers the 500 training pairs and
3,000 candidate pairs. When all dimensions apply, each candidate pair
requires nine dimension scores (G1, G2, F1, and Task Check on each side,
and F2 per pair), giving 27,000 dimension assessments per pass. Following
the wall-time accounting in \cref{app:diagnostics}, the measured
cumulative EoT-SFT plus RTPO time is 16.69 times the Stage-1 reference.
For a Stage-1 reference of approximately 0.80 hours, this ratio
corresponds to approximately 13.35 hours and 26.70 reserved GPU-hours
on two GPUs, including online-verifier waiting.

The controls of \cref{app:matched-controls} share the candidate count,
generation budget, and optimization settings. Their reward terms,
advantage organization, assignment of F2 scores, and generation context
follow the control definitions. Removing a scoring term reduces its verifier
calls while keeping the other coefficients fixed.

\section{Evaluation Protocols}
\label{app:evaluation}

\paragraph{Coverage and task scope.}
The main comparison uses HarborEval and OpenEval from FBA~\cite{fba},
together with harbor subsets of VRSBench~\cite{vrsbench},
RSVQA~\cite{rsvqa}, OmniEarth~\cite{omniearth}, and
VLRS-Bench~\cite{vlrsbench}. \Cref{tab:eval-inventory} lists the
archived evaluation inputs for both scene domains. The accompanying
\texttt{evaluation\_\allowbreak items.tsv} records the item IDs and source image
filenames. The HarborEval counts and item IDs in this inventory refer
to the earlier v1.2.1 split. Each two-date VLRS-Bench prompt presents both acquisitions
in order on a single image canvas.

\begin{table}[htbp]
  \centering
  \caption{Evaluation input lists. Images count distinct source image
  file paths within each list. Items count prompts, including report
  and caption prompts. Sensor counts below are counts of images.}
  \label{tab:eval-inventory}
  \small
  \setlength{\tabcolsep}{5pt}
  \begin{tabularx}{\textwidth}{@{}lrrX@{}}
    \toprule
    \textbf{Evaluation list} & \textbf{Images} & \textbf{Items} &
    \textbf{Input and task coverage} \\
    \midrule
    \multicolumn{4}{@{}l}{\textit{Harbor specialization}} \\
    HarborEval-lite v1.2.1 & 288 & 1,245 &
    Eight tracks over 184 RGB, 55 SAR, 32 panchromatic, and 17 near-infrared images. \\
    VRSBench harbor & 200 & 580 &
    RGB imagery with 430 scored VQA prompts and 150 auxiliary caption prompts. \\
    RSVQA harbor v3 & 78 & 719 &
    USGS test imagery with counting, presence, and spatial questions. \\
    OpenEval harbor & 100 & 100 &
    One open-ended prompt per image across 55 RGB, 25 SAR, 10 panchromatic,
    and 10 near-infrared images. \\
    OmniEarth harbor & 240 & 240 &
    Nine tasks covering perception, reasoning, and robustness, with segmentation tasks excluded. \\
    VLRS-Bench harbor & 176 & 132 &
    88 single-view questions and 44 questions using two acquisition dates. \\
    \midrule
    \multicolumn{4}{@{}l}{\textit{Urban specialization}} \\
    EarthVQA urban validation & 150 & 600 & Six question types on urban scenes. \\
    VRSBench urban VQA & 290 & 300 & Ten visual question types. \\
    RSVQA-HR urban test & 60 & 240 & 120 presence and 120 comparison questions. \\
    \bottomrule
  \end{tabularx}
\end{table}

\paragraph{Source-specific scoring.}
HarborEval's structured tasks cover recognition, functional zones,
relations, grid localization, observability, evidence judgment,
and scene rejection. Its source protocol uses accuracy for closed
choices, set F1 for multi-label answers, and soft grid F1 for
localization~\cite{fba}. In FBA, HarborEval's reporting track and
OpenEval share the six reporting criteria given below.

The public-source harbor protocol distinguishes VQA from detailed
captioning~\cite{fba,vrsbench,rsvqa}. Yes/no and numerical answers
use exact matching; other textual VQA answers use the source's
semantic-matching rules, while captioning compares generated and
reference descriptions for semantic consistency. A caption score
and a question-answering accuracy consequently retain different
interpretations even when expressed on the same reporting scale.

\paragraph{Closed and open-ended responses.}
Closed-ended tasks are scored against the answer key and accepted
variants for their question type, and model outputs are interpreted
according to the requested answer. Open-ended evaluation follows each
source protocol. HarborEval's reporting track uses a fixed image-grounded
judge separate from the training verifier~\cite{fba,mtbench}.

\paragraph{OpenEval scoring in \cref{tab:model_comparison}.}
In FBA's OpenEval evaluation, domain experts assessed open-ended
reports on 100 items spanning RGB, SAR, panchromatic, and near-infrared
imagery~\cite{fba}. The score sheet concealed model identity and
training route. Reviewers scored groundedness, object coverage, functional-zone
understanding, spatial relations, uncertainty and rejection, and
concision on a 0--5 scale. They recorded
dimension scores and brief notes and flagged unsupported claims, such
as an unverified port name or vessel identity. Borderline or
inconsistent ratings received an additional expert review, with the
adjudication status recorded on the score sheet. For GeoChat,
SkySenseGPT, VHM, SkyEyeGPT, and LHRS-Bot-Nova,
\cref{tab:model_comparison} reproduces the OpenEval scores that FBA
reports on a 0--100 scale in its Table~6.

For LLaVA-1.5, Qwen3.5-4B, EarthDial, RS-EoT, RSThinker, and G\&F,
MiMo-v2.6-Flash assessed one report per image on the 100-item
OpenEval set. The judge received the image, prompt, report, structured
reference, and item rubric. It scored 90 reports on seven dimensions
covering visual grounding, object and relation accuracy, functional
zones, uncertainty, hallucinations, modality, and report structure.
The ten non-harbor rejection items used five dimensions covering
hallucinations, uncertainty, grounding, objects, and task adherence.
For each report the judge returned dimension scores, a rationale, flags,
and a total on the 0--100 scale. We used the mean of the 100 reported
totals for each model. The G\&F totals sum to 8,207, giving the 82.07
in \cref{tab:model_comparison}. The same judge gave the Qwen3.5-4B
backbone 67.34.

\paragraph{Benchmark aggregation.}
Let $s_b$ denote the score on evaluation set $b$, expressed on the
0--100 reporting scale. The main summary is
\begin{equation}
\mathrm{Overall}=\frac16\sum_{b=1}^{6}s_b.
\label{eq:supp-overall}
\end{equation}
Each set receives equal weight regardless of its sample count.
Overall is a study-specific summary of the six reported metrics;
the individual scores retain their task-specific meaning.
The sets offer complementary assessments, with potential dependence
through shared capabilities or source imagery. Per-set results
provide the task-level context for interpreting the aggregate gain.

\paragraph{Supervision-scale interpretation.}
Data Scale in the main model comparison follows the adaptation
counts reported by each source. For G\&F, the annotation budget is
\sftcount{} demonstrations and 500 image pairs, with the image and card
counts of both stages given in \cref{app:data}.
These incremental annotations have a different accounting scope from
another model's full pretraining or instruction-tuning collection.
Limited supervision refers to annotation scale; the training-time
estimates are defined separately in \cref{app:diagnostics}.

\section{Experimental Settings}
\label{app:experimental-details}

\paragraph{Adaptation and inference.}
Harbor specialization uses Qwen3.5-4B~\cite{qwen35} with
LoRA~\cite{lora} and a frozen vision \mbox{encoder}; Stage~1 supplies the
common initialization for all harbor RL variants. Every evaluation
input contains one image and its question, and models use the input
formatting of their backbone.

\subsection{Joint-Input Generation Control}
\label{app:joint-control}

The joint-input control changes the context in which responses are
generated. One context contains both images and requests a description,
analysis, and answer for each side. In G\&F, each response instead
conditions only on its own image and the shared question.
Side-specific content extracted from the joint response is evaluated
with the same Gravity, Flight, and Task Check criteria. Later text
in the joint context may depend on earlier text and on either image;
this dependence is part of the generation-context comparison.

The comparison retains the source images, per-image preprocessing,
candidate-pair count, and verification weights. A complete image pair
is the accounting unit for generation. G\&F uses two single-image
calls, whereas joint-input generation includes both sides in one
context. The combined completion allowance covers both responses.
Under matched image--question exposure and generation limits,
attention cost, realized sequence length, and elapsed time may still
differ. The main table therefore reports measured training wall time
alongside performance, with its definition in \cref{app:diagnostics}.

All checkpoints are evaluated with the same single-image input
setting as G\&F.

\section{Training-Time and Verification Cost}
\label{app:diagnostics}

\paragraph{Reported cost.}
The relative costs in \cref{tab:ablation} report the training wall
time measured for each run under the optimized implementation,
including Stage~1 and online-verifier waiting. These times include
the request concurrency and synchronization of the optimized runtime
and are normalized to the Stage-1 SFT cost, $M_1=1.00\times$.
Each RL entry covers both initialization and subsequent training.
The accounting measures elapsed time; policy GPU-hours and API
expenditure are separate quantities.

\paragraph{Measurement.}
Each cost is the wall-clock time of the run that produced the
corresponding checkpoint, recorded end to end on the same two-GPU
machine. The record includes rollout generation, verifier requests, and
policy updates, and therefore reflects the concurrency and inter-rank
synchronization of each setting. Independent evaluation and manual
pauses fall outside the measured interval. One run is reported per
setting, so differences below a few percent should be read as
run-to-run variation rather than as systematic cost differences.

For each training unit, policy-gradient computation starts after its
valid scores and shared advantage are available. Generated tokens,
matching indices, and detached rollout-policy log probabilities are
retained through scoring and replay, so importance ratios use the
policy probabilities recorded for those responses.

\paragraph{Verification workload.}
Removing a scoring mechanism removes its calls while leaving the other
coefficients fixed. Random one-to-one matching requires $K$
paired-adaptation assessments per training unit, whereas an exhaustive
cross-product would require $K^2$; at $K=6$ this is six of 36
combinations. Total training time also includes generation, local
scoring, and policy updates.

\section{Urban Data and Evaluation}
\label{app:urban}

\paragraph{Data preparation.}
The urban instantiation uses 320 images at $1024\times1024$ from
EarthVL-GLOBAL in EarthVLSet~\cite{wang2026earthvl}, with 160
demonstrations, 60 training pairs, and 20 further pairs held out for
development. Related views remain within the same split, and adaptation
and evaluation use distinct image sets. Urban annotations follow the
same observation, analysis, and answer organization and retain the
harbor rubric dimensions and reward weights, while the scene content and
annotations are specific to urban data.

\paragraph{Training and evaluation.}
A separate urban Stage-1 model starts from the same base backbone and
initializes urban RTPO, and the comparison includes the base model,
urban EoT-SFT, and urban RTPO. The three checkpoints are evaluated on
fixed urban subsets of EarthVQA~\cite{earthvqa},
VRSBench~\cite{vrsbench}, and RSVQA HR~\cite{rsvqa}, each retaining its
task-specific scoring protocol under identical image--question inputs
and subset definitions. Inference remains single-image throughout.

\section{Token-Level Annotation and Statistics}
\label{app:token-analysis}

\paragraph{Example construction.}
The analysis in \cref{sec:token-evidence-organization} uses six
image--question cases randomly sampled from the evaluation set, each with
two required visual facts and a reference judgment. One response from the
Base checkpoint and one from G\&F are taken for the same input, giving
twelve responses; G\&F responses follow the
description--analysis--answer format, whereas Base responses use
unstructured prose.

\paragraph{Annotation and metrics.}
Task-bearing evidence phrases (E) and valid inference clauses (I) are
annotated with their parent sentence, and identical sentences receive
identical annotations across the two styles. Let $n_E$ and $n_I$ count
selected evidence and inference tokens, $n_D$ and $n_A$ count descriptive
and analytical content tokens, and $n_C$ count all content tokens. We
report evidence density $n_E/n_D$, inference density $n_I/n_A$, and
task-related density $(n_E+n_I)/n_C$, averaged over cases with equal
weights. Supported repetitions count toward density, while required-fact
coverage counts each fact once.

\paragraph{Statistics.}
Across the six cases, mean response length is 355.50 tokens for the Base
responses and 179.50 for G\&F (49.51\% shorter), with content lengths of
350.83 and 156.33. Evidence density is 21.53\% versus 38.48\% and
task-related density 22.35\% versus 39.91\%. Inference density, computed
on the five cases with analytical spans in both responses, is 24.82\%
versus 46.61\%. Required-fact coverage is 91.67\% versus 100.00\%, and
final answers agree with the reference judgment in 50.00\% of Base
responses and 100.00\% of G\&F responses. The G\&F responses state
evidence early and use it in the analysis, whereas the Base responses
distribute the same content across longer prose.

\end{document}